\documentclass{coowa}

\usepackage[toc,page,header]{appendix}
\usepackage{comment}

\usepackage{algorithm}
\usepackage{algpseudocode}

\usepackage{xspace}
\usepackage{amsmath}
\usepackage{amssymb}
\usepackage{amsfonts}
\usepackage{mathtools}

\usepackage{adjustbox}
\usepackage{accsupp}
\usepackage{transparent}

\usepackage{bbding}
\usepackage{wrapfig}

\usepackage{booktabs}
\usepackage{tabularx}
\usepackage{array}
\usepackage{colortbl}

\definecolor{PromptHeader}{HTML}{DCE8EE}
\definecolor{PromptSection}{HTML}{EAF1F4}
\definecolor{PromptLight}{HTML}{F6F9FA}

\newcolumntype{Y}{>{\raggedright\arraybackslash}X}

\crefname{algorithm}{Algorithm}{Algorithms}
\Crefname{algorithm}{Algorithm}{Algorithms}

\newcommand{\done}[1]{\textcolor{red}{[DONE]}}

\definecolor{ShortText}{RGB}{144,216,212}
\definecolor{LongText}{RGB}{128,200,195}
\definecolor{Visual}{RGB}{210,180,220}
\definecolor{bellow}{RGB}{229,181,89}
\definecolor{lightblue}{RGB}{230,240,250}

\newcommand{\bagel}{%
  \IfFileExists{sections/figures/bagel.png}{%
    \raisebox{-0.2ex}{\includegraphics[height=1em]{sections/figures/bagel.png}}%
  }{%
    \textcolor{bellow}{O}%
  }%
}

\title{RISE: Adaptive Imagination for World Action Models}

\author[1,2,*]{Hongbo Lu}
\author[1,3, *]{Liang Yao}
\author[1,*]{Chenghao He}
\author[1,*]{Hao Han}
\author[3]{Fan Liu}
\author[1,\dagger]{Wenlong Liao}
\author[1]{Tao He}
\author[1,\dagger,\ddagger]{Pai Peng}

\affiliation[1]{COWARobot Co. Ltd}
\affiliation[2]{Shanghai Jiao Tong University}
\affiliation[3]{Hohai University}

\contribution[*]{Equal Contribution}
\contribution[\dagger]{Corresponding Author}
\contribution[\ddagger]{Project Lead}

\abstract{
World Action Models (WAMs) improve planning by incorporating future world evolution into action generation, yet existing methods allocate a fixed imagination budget to every scene. We propose RISE (\textbf{R}efining \textbf{I}magination through \textbf{SE}lective Rollout), a system-level adaptive imagination framework that makes sequential \textsc{Roll}/\textsc{Stop} decisions according to the expected planning benefit of continued rollout. At each step, a Latent Evaluator estimates the risk revealed by the current prefix and how much planning could improve if imagination continues, while a Rollout Gate weighs this expected benefit against additional computation cost.
Since factual driving logs expose only one realized future, we further construct \textbf{CounterDrive}, a counterfactual dataset with diverse outcomes and risk levels, to enrich future dynamics and provide localized risk supervision. Each retained sample undergoes expert verification and annotation of trajectory validity, incident onset, and causal category, providing a reusable resource for safety-critical world-modeling research. Experiments on NAVSIM and nuScenes show that RISE achieves the best overall planning performance while reducing unnecessary rollout, with additional transfer results supporting its plug-in generality across WAM architectures.
}

\date{\today}

\checkdata[Project Page]{\url{https://cowarobot-ai.github.io/RISE/}}
\checkdata[Dataset]{\url{https://huggingface.co/datasets/COWARobot/CounterDrive}}
\checkdata[Codes]{\url{https://github.com/COOWAI/RISE}}
\correspondence{\email{volans.liao@cowarobot.com}, \email{pengpai@cowarobot.com}}

\begin{document}

\maketitle

%

\newpage
\tableofcontents
\newpage

\section{Introduction}

End-to-end autonomous driving~\cite{li2025hydra,shang2025drivedpo} is moving beyond direct trajectory regression toward models that reason about the future before acting~\cite{chen2025vl,maes2026leworldmodel}. World Action Models (WAMs)~\cite{shen2026WAMsurvey,wang2026WAMsurvey2} incorporate predicted world evolution into planning, allowing policies to evaluate actions through their consequences rather than the current observation alone. This capability is particularly relevant to driving, where similar visual contexts can lead to different outcomes as the ego vehicle and surrounding agents interact. Consequently, recent methods~\cite{wang2026drive,yang2026worldrft} employ latent future prediction, action-conditioned dynamics, or world-model-based policy learning to improve planning under interaction and uncertainty.

\begin{figure}[t]
    \centering
    \includegraphics[width=0.5\linewidth]{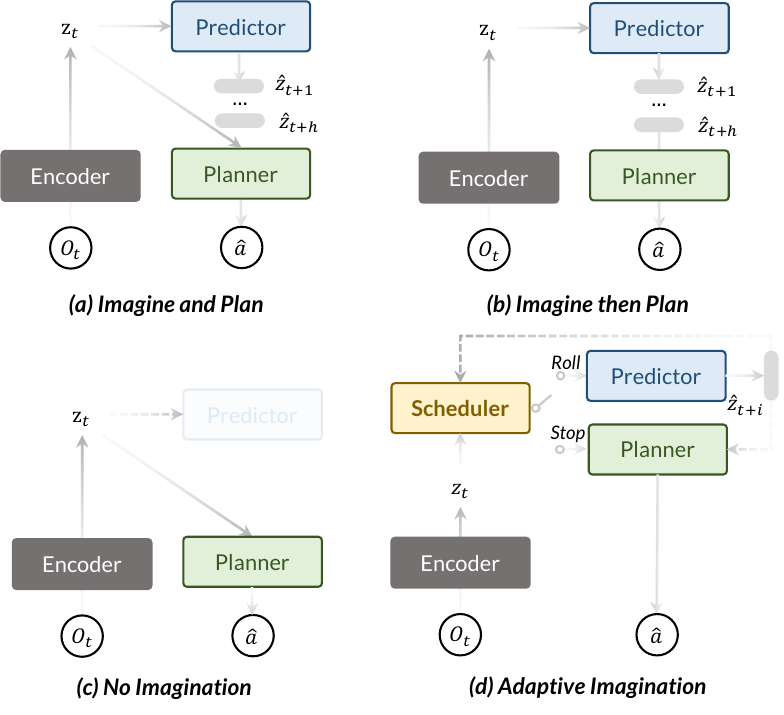}
    \caption{Comparison of various imagination strategies in WAMs. (a) \emph{Imagine and Plan} jointly uses observed and predicted latents for planning, (b) \emph{Imagine then Plan} performs future rollout before action generation, and (c) \emph{No Imagination} plans directly from the observation. (d) RISE introduces a lightweight Scheduler that adaptively routes computation to further prediction or final planning.}
    \label{fig:teaser}
\end{figure}

However, future imagination adds inference cost because latent futures must be generated before an action can be produced. As illustrated in Fig.~\ref{fig:teaser}(a)--(c), existing WAMs~\cite{wang2026latentwam,yang2026dreamera,yuan2026fast,lu2026dawn} organize this computation differently. Shared-backbone models perform prediction and planning from a common representation through an \emph{imagine-and-plan} design. Cascaded models complete a future rollout before passing it to the Planner through an \emph{imagine-then-plan} design. Other approaches remove test-time imagination and plan directly from the observation. Despite these differences, their inference schedules are specified globally. They do not reassess after each partial rollout whether the current prefix is sufficient for planning or continued prediction is likely to improve the final decision.

We refer to this missing decision signal as \emph{Future Planning Gain}. At rollout step, its profile measures the change in planning score associated with each valid continuation relative to stopping with the current prefix. By expressing continuation utility in planning score, it directly connects additional rollout to downstream planning quality. Because each predicted latent updates the current prefix and the available future evidence, the gain may change throughout rollout. Therefore, it should be re-estimated after each step and used as a sequential stopping signal rather than to select a final rollout depth in advance.

This perspective captures the varying utility of imagination without dividing scenes into predefined easy and difficult categories. The observation or an early prefix may already support a stable trajectory, so continuing offers little gain and only increases latency. Alternatively, the current prefix may leave an interaction unresolved, and an additional predicted step may change the preferred plan. In either case, the decision is local: the model should \textsc{Roll} when the expected improvement justifies its cost and \textsc{Stop} otherwise. Repeating this decision produces a horizon ranging from zero rollout to the dataset-specific maximum.

Building on this principle, we introduce \textsc{RISE} (\textbf{R}efining \textbf{I}magination through \textbf{SE}lective Rollout), the plug-in adaptive imagination framework illustrated in Fig.~\ref{fig:teaser}(d). RISE augments an Encoder--Predictor--Planner WAM with a lightweight Scheduler composed of a Latent Evaluator and a Rollout Gate. At each step, the Evaluator predicts a Risk Profile summarizing the planning-relevant risk already revealed and a Future Planning Gain Profile describing the improvement available through continued rollout. The Gate weighs the predicted gain against computation cost and makes the current binary decision. Under \textsc{Roll}, the Predictor appends one future latent and the Scheduler evaluates the updated prefix again. Under \textsc{Stop}, the selected prefix is passed once to the Planner. The Gate does not predict the final horizon directly. Therefore, the effective horizon emerges from repeated \textsc{Roll}/\textsc{Stop} decisions.

The two Evaluator signals require different supervision. Planning scores obtained by stopping at every valid prefix provide direct Future Planning Gain targets: the score difference between a continued and current prefix measures the realized benefit of continuation. Risk learning faces a different limitation because factual logs record only the future that occurred and leave plausible safety-critical alternatives unseen. We construct \textsc{CounterDrive} to augment selected contexts with diverse counterfactual outcomes and risk levels. Accepted clips enrich future-latent prediction, while verified factual--counterfactual incident pairs provide temporally localized Risk Profile supervision. Consequently, all-horizon planning outcomes teach how planning changes under continuation, while CounterDrive improves sensitivity to risks revealed along that process.

Extensive experiments on NAVSIM and nuScenes validate the effectiveness of \textsc{RISE}. It achieves 91.5 PDMS and 90.8 EPDMS on NAVSIM v1 and v2, respectively, while obtaining state-of-the-art trajectory accuracy and collision performance on nuScenes. Further analyses show that CounterDrive improves hazard discrimination and that the Scheduler transfers to another architecture without modifying its Predictor or Planner, supporting both adaptive imagination and architectural generality. Our contributions are summarized as follows:
\begin{itemize}
    \item We propose \textbf{RISE}, a plug-in adaptive imagination framework for WAMs that uses Future Planning Gain to make sequential \textsc{Roll}/\textsc{Stop} decisions, yielding a scene-dependent rollout horizon while balancing planning quality and inference cost.
    
    \item We construct \textbf{CounterDrive}, a counterfactual driving dataset with diverse interaction outcomes and risk levels, supporting both RISE training and safety-critical world modeling research.
    
    \item Experiments on driving benchmarks show that RISE achieves state-of-the-art planning performance with lower rollout cost, while its transfer to other WAM architecture supports plug-in generality.
\end{itemize}
\section{Related Works}


\subsection{World Action Models}

World action models (WAMs) extend predictive world modeling~\cite{ha2018world,zhou2024dino} with executable action generation. Rather than directly mapping observations to actions, they use future prediction to learn physical dynamics and state--action correspondences. Unified Video Action Model~\cite{li2025unified} learns a shared video--action latent space for forward prediction, inverse dynamics, and real-time control. Cosmos Policy~\cite{kim2026cosmos} adapts pretrained video diffusion models to jointly predict future states, actions, and values. LingBot-VA~\cite{li2026causal}, DreamZero~\cite{ye2026worldaction}, MotuBrain~\cite{xiang2026motubrain}, and WLA-0~\cite{yang2026worldlanguageaction} further scale WAMs toward language-conditioned, long-horizon, and cross-embodiment control. However, most existing WAMs treat test-time imagination as a fixed architectural choice.

\subsection{Autonomous Driving World Models}

Driving world models predict scene evolution from historical observations and ego actions~\cite{feng2025survey}. ReSim~\cite{yang2026resim} further studies reliable simulation for autonomous driving. GAIA-1~\cite{hu2023gaia} and DriveDreamer~\cite{wang2023drivedreamer} generate controllable futures from video, text, and action conditions. OccWorld~\cite{zheng2024occworld} models future 3D occupancy, while HERMES~\cite{zhou2025hermes} and UniFuture~\cite{liang2025seeing} couple generation with scene understanding or geometric perception. DriveDreamer4D~\cite{zhao2025drivedreamer4d} uses generated futures to improve 4D scene representations. More recent methods emphasize planning-oriented latent dynamics: Epona~\cite{zhang2025epona} integrates autoregressive diffusion with trajectory planning, whereas Latent-WAM~\cite{wang2026latentwam}, DriveFuture~\cite{hong2026drivefuture}, and DreamerAD~\cite{yang2026dreamera} learn compact future states for end-to-end planning or reinforcement learning. Despite their different representations, these methods generally use a fixed imagination strategy, whereas RISE allocates future reasoning according to scene-dependent planning utility.
\section{CounterDrive}

Factual driving logs contain only realized futures and provide limited safety-critical alternatives. Therefore, we construct \textsc{CounterDrive} from selected NAVSIM and nuScenes scenes. Each retained counterfactual clip is associated with its factual source, forming pairs within the selected subset rather than covering the complete datasets.

\subsection{Counterfactual Video Generation}
For each selected source key frame, the prompt combines a fixed instruction, a scene description, and an incident description specifying the event location and involved object. Prompts additionally constrain the camera viewpoint, road geometry, background, and initial traffic configuration so that the generated clip remains anchored to the source scene. Conditioned on the key frame and prompt, Wan~2.7 generates a 10-second video at 1080p resolution, which is sampled at 2\,Hz into 20 frames.

\subsection{Trajectory Annotation}
We apply OpenVO~\cite{nguyen2026openvo} to recover the frame-wise ego poses:
\begin{equation}
    \tilde{\tau}_{i,t}
    =
    (\tilde{p}^{x}_{i,t},
     \tilde{p}^{y}_{i,t},
     \tilde{\psi}_{i,t}).
\end{equation}
The corresponding ego-motion actions are computed from adjacent poses:
\begin{equation}
    \tilde{a}_{i,t}
    =
    \mathcal{A}
    (\tilde{\tau}_{i,t},
     \tilde{\tau}_{i,t+1}).
\end{equation}

\subsection{Human Verification and Annotation}
Annotators verify ego-motion consistency, identify the first incident frame, mark generation distortions, and categorize each clip as normal, caused by non-ego behavior, or caused by ego behavior. A recovered trajectory is marked invalid when it disagrees with the visually observed ego motion. For ego-caused incidents, annotators also record recommended avoidance or stopping actions as clip-level metadata. Clips with severe distortion or unreliable motion are removed. Accepted clips supervise future prediction, while verified incident pairs provide temporally localized risk-ranking supervision.

After filtering, CounterDrive contains 2,432/511 training/test clips from nuScenes and 5,013/1,000 from NAVSIM. Unpaired factual samples remain available for standard training objectives. Additional construction and annotation details are provided in the Appendix.

\section{RISE}

\begin{figure*}[t]
    \centering
    \includegraphics[width=\linewidth]{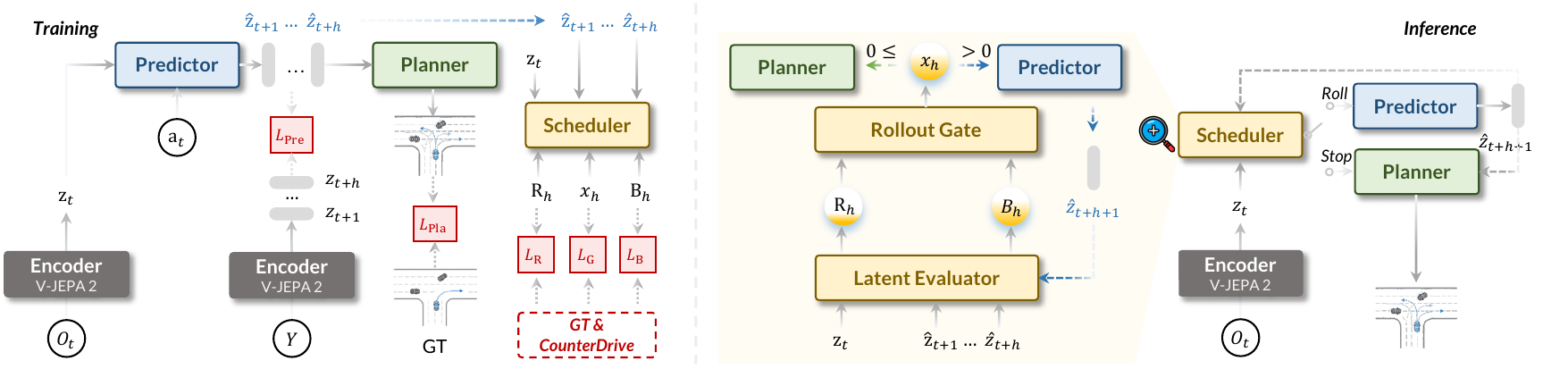}
    \caption{Overview of RISE. The WAM learns future prediction and variable-prefix planning. The Latent Evaluator predicts a Risk Profile $R_h$ and a Future Planning Gain Profile $B_h$, supervised by counterfactual risks and all-horizon planning outcomes. The Rollout Gate balances the predicted gain against computation cost, routing $x_h>0$ to further rollout and $x_h\leq0$ to planning.}
    \label{fig:overview}
\end{figure*}
We propose \textsc{RISE} (\textbf{R}efining \textbf{I}magination through \textbf{SE}lective Rollout), a plug-in adaptive imagination framework for World Action Models. As shown in Fig.~\ref{fig:overview}, RISE augments an Encoder--Predictor--Planner WAM with a lightweight Scheduler composed of a Latent Evaluator and a Rollout Gate. The Evaluator estimates the planning risk revealed by the current future prefix and the potential gain of deeper rollout, while the Gate determines whether additional imagination is worth its computation cost.

\subsection{Problem Formulation}
\label{sec:problem_formulation}

Let $c$ denote the current driving context, $\tau_{1:P}$ the future trajectory over planning horizon $P$, and $z_{1:H}$ the future world representation over a maximum rollout horizon $H$. A standard World Action Model uses a fixed rollout depth for every scene:
\begin{equation}
p(z_{1:H},\tau_{1:P}\mid c)
=
p(z_{1:H}\mid c)\,
p(\tau_{1:P}\mid c,z_{1:H}).
\end{equation}
RISE instead makes the effective rollout depth scene-dependent. Given an imagined prefix
$\hat Z_h=\hat z_{1:h}$, where $\hat Z_0=\emptyset$, the Scheduler determines whether the current prefix is sufficient for planning:
\begin{equation}
\begin{aligned}
d_h
&=
\mathcal{S}(\hat Z_h;c,\lambda),\\
K(c;\lambda)
&=
\min\left(
\{h\mid d_h=\textsc{Stop}\}
\cup
\{H\}
\right),
\end{aligned}
\label{eq:adaptive_horizon}
\end{equation}
where $\lambda$ controls the preference for computation cost. 

Our RISE models
\begin{equation}
p(z_{1:K(c;\lambda)},\tau_{1:P}\mid c)
=
p(z_{1:K(c;\lambda)}\mid c)\,
p(\tau_{1:P}\mid c,z_{1:K(c;\lambda)}).
\end{equation}
Here, $K(c;\lambda)\in\{0,\ldots,H\}$ is selected independently for each scene. When $K=0$, the model plans directly from the observed context; when $K=H$, it recovers the full-rollout behavior of a standard WAM.


\subsection{World Action Model Architecture}

\subsubsection{Encoder}
We use a frozen V-JEPA~2~\cite{assran2025v} image encoder with a ViT-L backbone to encode front-camera observations into latent tokens. Four input frames are resized to $256\times512$ and processed using $16\times16$ patches and non-overlapping two-frame tubelets. The Encoder produces two observed latent time steps, each containing 512 spatial tokens of dimension 1024.

\subsubsection{Predictor}
We adopt a frame-causal ego motion conditioned Transformer with 12 layers, a hidden dimension of 384, and 12 attention heads. Starting from the observed latent tokens, the Predictor autoregressively generates up to $H_{\mathcal D}$ future latent steps, each corresponding to two raw video frames. Hereafter, $H_{\mathcal D}$ denotes the dataset-specific realization of the maximum horizon $H$. Ego-motion conditions are represented by relative pose changes computed between adjacent ego poses. The Predictor uses no route command, additional state input, or camera extrinsic.

\subsubsection{Planner}
We employ a 12-layer diffusion Transformer with a hidden dimension of 384 and 12 attention heads. The Planner conditions on the observed tokens and the selected future prefix, distinguished by learned source embeddings, together with the observed ego-motion history and current ego kinematics. It generates six candidate trajectories using a variance-preserving diffusion process with 20 DPM-Solver++~\cite{lu2025dpm} sampling steps. Each candidate contains eight future poses at $0.5$\,s intervals, represented as $(x,y,\cos\psi,\sin\psi)$. Candidate selection and dataset-specific scoring details are provided in the Appendix.

\subsection{Adaptive Rollout Scheduler}
\label{sec:scheduler}

At rollout depth $h$, the Scheduler decides whether to stop with the currently available future prefix or generate one additional latent step. It is guided by the predicted Future Planning Gain of continuing rollout and computation cost. 

\subsubsection{Latent Evaluator}
Given the observed latent $z_t$ and imagined prefix $\hat Z_h$, the Latent Evaluator predicts a Risk Profile and a Future Planning Gain Profile:
\begin{equation}
\begin{aligned}
(R_h,B_h)
&=
\operatorname{Eval}_{\theta_E}
(z_t,\hat Z_h),\\
R_h
&=
[r_1,\ldots,r_h],\\
B_h
&=
[b_{h\rightarrow h+1},\ldots,
 b_{h\rightarrow H_{\mathcal D}}],
\end{aligned}
\label{eq:latent_evaluator}
\end{equation}
where $r_k$ estimates the trajectory risk associated with the prefix ending at depth $k$, while $b_{h\rightarrow j}$ predicts the Future Planning Gain of continuing from depth $h$ to $j$, measured relative to stopping with the current prefix. Thus, $R_h$ summarizes the risk already exposed, whereas $B_h$ summarizes the potential planning gains available from continued rollout. 

\subsubsection{Rollout Gate}
For $0\leq h<H_{\mathcal D}$, we construct
\begin{equation}
\begin{aligned}
e_h
&=
[\rho(z_t),\rho(\hat Z_h)],\\
\xi_h
&=
[\overline{R}_h,
 \overline{B}_h,
 h/H_{\mathcal D},
 c_h,
 \Delta c_{h+1}],
\end{aligned}
\label{eq:gate_input}
\end{equation}
where $\rho(\cdot)$ denotes token pooling,
$\rho(\emptyset)$ is a learned empty-prefix embedding,
$\overline{R}_h$ and $\overline{B}_h$ are zero-padded to length $H_{\mathcal D}$, $c_h$ is the cumulative cost, and
$\Delta c_{h+1}=c_{h+1}-c_h$. 
Given computation preference $\lambda$, the Gate predicts
\begin{equation}
x_h
=
G_{\theta_G}(e_h,\xi_h,\lambda).
\label{eq:gate_score}
\end{equation}
The sign of $x_h$ indicates whether the predicted planning gain available from continued rollout justifies its additional cost. The Scheduler applies
\begin{equation}
d_h=
\begin{cases}
\textsc{Roll}, & x_h>0\ \text{and}\ h<H_{\mathcal D},\\
\textsc{Stop}, & \text{otherwise}.
\end{cases}
\label{eq:rollout_decision}
\end{equation}
For \textsc{Roll}, the Predictor appends one future latent step and the Scheduler evaluates the extended prefix. For \textsc{Stop}, the selected prefix is risk-refined when $h>0$ and passed once to the Planner. Refinement is skipped when $h=0$, and rollout terminates at $H_{\mathcal D}$ by construction.

\subsection{Three-Stage Training}
\label{sec:training}

RISE is trained in three stages. Stage I trains the Predictor and an initial variable-prefix Planner. Stage II learns planning-oriented risk and Future Planning Gain. Stage III converts these signals into a cost-aware stopping policy.

\subsubsection{Stage I: Predictor and Initial Planner}
The Predictor recursively generates the maximum dataset-specific prefix
$\hat Z_{H_{\mathcal D}}$. It is trained on real and accepted CounterDrive sequences using teacher-encoded future latents:
\begin{equation}
\mathcal{L}_{\mathrm{Pre}}
=
\sum_{k=1}^{H_{\mathcal D}}
\ell_z
\left(
\hat z_{t+k},
\operatorname{sg}(z_{t+k})
\right),
\label{eq:prediction_loss}
\end{equation}
where $z_{t+k}$ is the target latent, $\operatorname{sg}(\cdot)$ denotes the stop-gradient. For CounterDrive sequences, the ego-motion conditions are computed from adjacent recovered poses. We then train an initial Planner $\Pi_0$ using real trajectory supervision:
\begin{equation}
\mathcal{L}_{\mathrm{Pla}}^{0}
=
\ell_{\tau}
\left(
\Pi_0(z_t,\hat Z_h),
\tau^{*}
\right),
h\sim\{0,\ldots,H_{\mathcal D}\}.
\label{eq:initial_planner_loss}
\end{equation}
The Encoder, Predictor, and $\Pi_0$ are frozen when constructing the subsequent training targets.

\subsubsection{Stage II: Latent Evaluator and Guided Planner}
For each real prefix, $\Pi_0$ produces candidate trajectories whose planning risk is computed by a fixed geometry-based evaluator:
\begin{equation}
\begin{aligned}
r_k^{*}
&=
\mathcal{E}_{\mathrm{risk}}
\left(
\Pi_0(z_t,\hat Z_k)
\right),\\
\mathcal{L}_{\mathrm{real}}
&=
\sum_{k=1}^{H_{\mathcal D}}
\ell_{\mathrm{Huber}}
\left(
r_k,r_k^{*}
\right).
\end{aligned}
\label{eq:risk_supervision}
\end{equation}
The evaluator includes the common candidate-selection rule and geometry-based risk terms, whose definitions are provided in the Appendix.

CounterDrive contributes paired supervision only when a verified factual--counterfactual pair is available. Let $\mathcal I_{\mathrm{pair}}$ denote the verified pairs in which the counterfactual future contains an annotated incident and is ranked as riskier than its factual source. For a pair $i$, the annotated incident frame is mapped to latent step $k_i^{\mathrm{inc}}$, and we define
\begin{equation}
\begin{aligned}
m_{i,k}
&=
\mathbb{I}
\left[
k\geq k_i^{\mathrm{inc}}
\right],\\
\mathcal{L}_{\mathrm{rank}}
&=
\sum_{i\in\mathcal I_{\mathrm{pair}}}
\sum_{k=1}^{H_{\mathcal D}}
m_{i,k}
\left[
\gamma
+
r_{i,k}^{\mathrm{fac}}
-
r_{i,k}^{\mathrm{cf}}
\right]_{+}.
\end{aligned}
\label{eq:risk_ranking}
\end{equation}
Real samples without a paired counterfactual clip are excluded only from this ranking term. The complete Risk Profile objective is
\begin{equation}
\mathcal{L}_{R}
=
\mathcal{L}_{\mathrm{real}}
+
\beta_{\mathrm{cf}}\mathcal{L}_{\mathrm{rank}}
+
\beta_{\mathrm{loc}}\mathcal{L}_{\mathrm{loc}},
\label{eq:risk_loss}
\end{equation}
where $\mathcal{L}_{\mathrm{loc}}$ uses the annotated incident onset for temporally localized risk calibration. Its detailed construction and loss weights are provided in the Appendix.

After learning the Risk Profile, we refine each non-empty prefix using a small norm-constrained update:
\begin{equation}
\tilde Z_h^{m+1}
=
\tilde Z_h^{m}
-
\eta
\nabla_{\tilde Z_h^{m}}
\sum_{k=1}^{h}r_k,
\tilde Z_h^{0}=\hat Z_h.
\label{eq:risk_guidance}
\end{equation}
For $h=0$, we define $\tilde Z_0=\emptyset$ and skip refinement. The final Planner $\Pi_1$ is trained at all valid rollout depths:
\begin{equation}
\mathcal{L}_{\mathrm{Pla}}^{1}
=
\ell_{\tau}
\left(
\Pi_1(z_t,\tilde Z_h),
\tau^{*}
\right),
h\sim\{0,\ldots,H_{\mathcal D}\}.
\label{eq:guided_planner_loss}
\end{equation}

We evaluate $\Pi_1$ at every rollout depth and denote its dataset-specific planning score by
\begin{equation}
q_h
=
\mathcal{E}_{\mathrm{plan}}
\left(
\Pi_1(z_t,\tilde Z_h),
\tau^{*}
\right).
\end{equation}
The Future Planning Gain target is
\begin{equation}
\begin{aligned}
B_h^{*}
&=
[q_{h+1}-q_h,\ldots,
 q_{H_{\mathcal D}}-q_h],\\
\mathcal{L}_{B}
&=
\sum_{h=0}^{H_{\mathcal D}-1}
\ell_B
\left(
B_h,B_h^{*}
\right).
\end{aligned}
\label{eq:gain_profile}
\end{equation}
The loss details are in the Appendix.

\subsubsection{Stage III: Rollout Gate}
Finally, we freeze other modules and train the Rollout Gate as a cost-aware stopping policy. For each real sample, we enumerate all valid rollout depths and record $q_h$ and cumulative cost $c_h$. For a computation preference $\lambda$, the best remaining cost-adjusted gain is
\begin{equation}
g_h^{*}
=
\max_{j\in\{h+1,\ldots,H_{\mathcal D}\}}
\left[
(q_j-q_h)
-
\lambda(c_j-c_h)
\right].
\label{eq:cost_adjusted_gain}
\end{equation}
The continuation target and Gate objective are
\begin{equation}
\begin{aligned}
y_h^{*}
&=
\mathbb{I}
\left[
g_h^{*}>0
\right],\\
\mathcal{L}_{G}
&=
\sum_{h=0}^{H_{\mathcal D}-1}
\ell_{\mathrm{BCE}}
\left(
\sigma(x_h),
y_h^{*}
\right).
\end{aligned}
\label{eq:gate_loss}
\end{equation}
At the maximum depth, rollout stops by construction. Full-horizon enumeration is used only to construct training targets. 
\section{Experiments}

\begin{table}[t]
\centering
\setlength{\tabcolsep}{10pt}
\begin{tabular}{l|cccc|cccc}
\toprule
\multirow{2}{*}{Method} & \multicolumn{4}{c|}{L2 (m)$\downarrow$} & \multicolumn{4}{c}{Collision Rate$\downarrow$} \\
\cline{2-9}
                        & 1s    & 2s   & 3s   & Avg. & 1s      & 2s     & 3s     & Avg.   \\
\hline                        
BEV-Planner             & 0.30  & 0.52 & 0.83 & 0.55 & 0.10    & 0.37   & 1.30   & 0.59   \\
LAW                     & 0.26  & 0.57 & 1.01 & 0.61 & 0.14    & 0.21   & 0.54   & 0.30   \\
World4Drive             & 0.23  & 0.47 & 0.81 & 0.50 & 0.02    & 0.12   & 0.33   & 0.16   \\
WorldRFT                & 0.21  & 0.44 & 0.76 & 0.47 & 0.10    & 0.11   & 0.23   & 0.15   \\
DAWN                    & 0.17  & 0.31 & 0.52 & 0.33 & 0.00    & \textbf{0.10}   & 0.23   & 0.11   \\
\hline
RISE                    &       \textbf{0.16}&      \textbf{0.29}&      \textbf{0.49}&      \textbf{0.31}&         \textbf{0.00}&        0.11&        \textbf{0.20}&       \textbf{0.10}\\
\bottomrule
\end{tabular}
\caption{Performance on nuScenes.}
\label{main:nuScnens}
\end{table}

\begin{table}[t]
\centering
\setlength{\tabcolsep}{10pt}
\begin{tabular}{l|cccccc}
\toprule
Method             & NC$\uparrow$  & DAC$\uparrow$ & EP$\uparrow$  & C$\uparrow$   & TTC$\uparrow$ & PDMS$\uparrow$ \\
\hline
DrivingGPT         & 98.9 & 90.7 & 79.7 & 95.6 & 94.9 & 82.4  \\
LAW                & 97.4 & 93.3 & 78.8 & 100  & 91.9 & 83.8  \\
World4Drive        & 97.4 & 94.3 & 79.9 & 100  & 92.8 & 85.1  \\
Epona              & 97.9 & 95.1 & 80.4 & 99.9 & 93.8 & 86.2  \\
DriveVLA-W0        & 98.4 & 95.3 & 80.9 & 100  & 95.2 & 87.2  \\
PWM                & 98.6 & 95.9 & 81.8 & 100  & 95.4 & 88.1  \\
DreamerAD          & 98.0 & 97.2 & 83.1 & 100  & 94.3 & 88.7  \\
DriveLaW           & 99.0 & 97.1 & 81.3 & 100  & 96.7 & 89.1  \\
Drive-JEPA         & 98.7 & 96.2 & 82.9 & 100  & 95.5 & 89.0  \\
DAWN               & 98.7 & 95.9 & 84.3 & 100  & 96.0 & 89.1  \\
EponaV2            & 98.6 & 97.9 & 84.8 & 100  & 95.7 & 90.4  \\
DriveFuture        & 98.8 & \textbf{99.1} & 95.4 & 100  & 84.2 & 90.7  \\
\hline
RISE               &      \textbf{99.1}&      97.7&      \textbf{98.3}&      100&      \textbf{98.6}&       \textbf{91.5}\\
\bottomrule
\end{tabular}
\caption{Performance on NAVSIMv1.}
\label{main:navsimv1}
\end{table}

\subsection{Implementation Details}
All experiments are implemented in PyTorch using bfloat16 precision. The training is conducted on 8 $\times$ NVIDIA A100 GPUs with a per-GPU batch size of 4. Videos are sampled at 2\,Hz and resized to $256\times512$. Each clip contains four observed frames and a dataset-specific future horizon. With two-frame tubelets, we set $H_{\mathrm{NAVSIM}}=4$ and $H_{\mathrm{nuScenes}}=3$. We train $\Pi_0$ for 20 epochs with a learning rate of $2\times10^{-5}$ and $\Pi_1$ for 50 epochs with $5\times10^{-5}$ using AdamW. The scheduler is trained for 50 epochs with a learning rate of $10^{-3}$. Latent guidance performs two gradient steps with a step size of $0.05$ and a maximum update norm of $0.25$. We use normalized rollout costs $c_h=h$ and evaluate $\lambda\in\{0,0.001,0.005,0.01,0.05\}$, with $0.005$ as the default. 

\subsection{Main Results}

\begin{table*}[t]
\centering
\begin{tabular}{l|cccccccccc}
\toprule
Method      & NC$\uparrow$  & DAC$\uparrow$ & DDC$\uparrow$ & TL$\uparrow$  & EP$\uparrow$  & TTC$\uparrow$ & LK$\uparrow$  & HC$\uparrow$  & EC$\uparrow$  & EPDMS$\uparrow$ \\
\hline
DAWN~\cite{lu2026dawn}        & 97.3 & 92.0 & 99.1 & 99.7 & 87.4 & 96.6 & 96   & 98.3 & 85.5 & 83.2   \\
DreamerAD~\cite{yang2026dreamera}   & 98.0 & 97.2 & 99.5 & 99.8 & 87.8 & 97.4 & 97.5 & 98.3 & 72.4 & 85.1   \\
DriveLaW~\cite{xia2026drivelaw}   & 98.7 & 96.9 & 99.6 & 99.8 & 87.5 & 98.3 & 97.6 & 98.4 & 77.4 & 88.6   \\
EponaV2~\cite{xu2026eponav2}     & 98.5 & 97.4 & 99.5 & 99.9 & \textbf{87.9} & 98.1 & 97.7 & 98.2 & 77.4 & 88.9   \\
Latent-WAM~\cite{wang2026latentwam}  & 98.1 & 97.3 & 99.6 & 99.8 & 87.7 & 97.3 & 97.6 & 98.1 & 87.3 & 89.3   \\
DriveFuture~\cite{hong2026drivefuture} & 98.8 & 99.1 & 99.6 & 99.9 & 86.6 & 98.4 & 96.4 & 98.3 & 74.8 & 89.9   \\
\hline
\textbf{RISE}        & \textbf{99.1}&  \textbf{97.7}&      \textbf{99.7}&      \textbf{99.9}&      87.8&      \textbf{98.7}&      \textbf{98.0}&      \textbf{98.4}&      \textbf{87.4}&        \textbf{90.8}\\
\bottomrule
\end{tabular}
\caption{Performance on NAVSIMv2.}
\label{main:navsimv2}
\end{table*}

We compare RISE with representative driving WAMs in the perception-free setting, including DrivingGPT~\cite{chen2024drivinggpt}, LAW~\cite{li2024enhancing}, World4Drive~\cite{zheng2025world4drive}, Epona~\cite{zhang2025epona}, DriveVLA-W0~\cite{li2025drivevlaw0}, PWM~\cite{zhao2025forecasting}, DreamerAD~\cite{yang2026dreamera}, DriveLaW~\cite{xia2026drivelaw}, Drive-JEPA~\cite{wang2026drive}, DAWN~\cite{lu2026dawn}, EponaV2~\cite{xu2026eponav2}, DriveFuture~\cite{hong2026drivefuture}, and Latent-WAM~\cite{wang2026latentwam}. As shown in Tables~\ref{main:nuScnens},~\ref{main:navsimv1} and~\ref{main:navsimv2}, RISE achieves the best overall planning performance on both NAVSIM V1 and V2, reaching 91.5 PDMS and 90.8 EPDMS, respectively. It surpasses the strongest baselines by 0.8 points on V1 and 0.9 points on V2. On NAVSIM V1, RISE improves the previous best EP and TTC by 2.9 and 1.9 points, respectively. On NAVSIM V2, it ranks first or ties for first on seven of the nine component metrics, demonstrating strong performance across safety, compliance, and planning quality. RISE also establishes state-of-the-art results on nuScenes, achieving the lowest average L2 error of 0.31\,m and collision rate of 0.10. These results demonstrate that adaptive rollout improves planning across evaluation settings.

\subsection{Ablation Studies}

\subsubsection{Ablation on Key Components}

\begin{wraptable}[9]{r}{0.46\textwidth}
\vspace{-4mm}
\setlength{\tabcolsep}{4pt}
\centering
\begin{tabular}{cc|cc}
\toprule
Scheduler  & CounterDrive   & EPDMS$\uparrow$   &PDMS$\uparrow$\\
\hline
           &                &         88.9& 89.7\\
           &   \checkmark   &         89.8& 90.5\\
\checkmark &                &         90.4& 91.2\\
\checkmark & \checkmark     &         \textbf{90.8}& \textbf{91.5}\\
\bottomrule
\end{tabular}
\caption{Ablation of the Scheduler and CounterDrive on NAVSIM. Both components improve performance. }
\label{ablation1}
\end{wraptable}

Table~\ref{ablation1} isolates the Scheduler and CounterDrive under the same backbone and training settings. CounterDrive alone raises EPDMS/PDMS from 88.9/89.7 to 89.8/90.5, indicating that counterfactual futures provide useful supervision for future representation learning. The Scheduler alone reaches 90.4/91.2, confirming that scene-dependent rollout allocation independently improves planning. Combining both yields the best results of 90.8/91.5, showing their complementary roles: CounterDrive enriches future supervision, while the Scheduler allocates additional imagination where it remains useful.

\subsubsection{Ablation on Adaptive Rollout Method}

\begin{wraptable}[9]{r}{0.6\textwidth}
\vspace{-4mm}
\setlength{\tabcolsep}{3pt}
\centering
\begin{tabular}{l|cc|c}
\toprule
Method        & \# Avg. Rollout & Latency (ms) & EPDMS$\uparrow$ \\
\hline
Random Stop   &     2.03&      \textbf{264.075}&  89.5\\
Latent Margin &     2.98&     308.532&    89.7\\
Scheduler     &     2.40&    287.429&      \textbf{90.8}\\
\bottomrule
\end{tabular}
\caption{Ablations on different adaptive rollout strategies. Random Stop stops at a random horizon from 0 to 4, while Latent Margin stops when consecutive latents converge.}
\label{ablation2}
\end{wraptable}

Table~\ref{ablation2} compares the Scheduler with Random Stop and Latent Margin. Random Stop assigns each sample a reproducible horizon from $\{0,\ldots,4\}$, whereas Latent Margin stops when consecutive latents converge. Random Stop provides the lowest latency of 264.075\,ms but reaches only 89.5 EPDMS. Latent Margin averages 2.98 rollouts and incurs 308.532\,ms latency, yet obtains only 89.7 EPDMS. In contrast, the Scheduler achieves 90.8 EPDMS with 2.40 rollouts and 287.429\,ms latency. It shows that effective stopping should reflect the planning gain rather than random allocation or latent convergence alone.


\subsection{Further Analysis}

\subsubsection{Is Adaptive Rollout Necessary?}

\begin{wrapfigure}[15]{r}{0.40\columnwidth}
\vspace{-0.35cm}
    \centering
    \includegraphics[width=0.4\columnwidth]{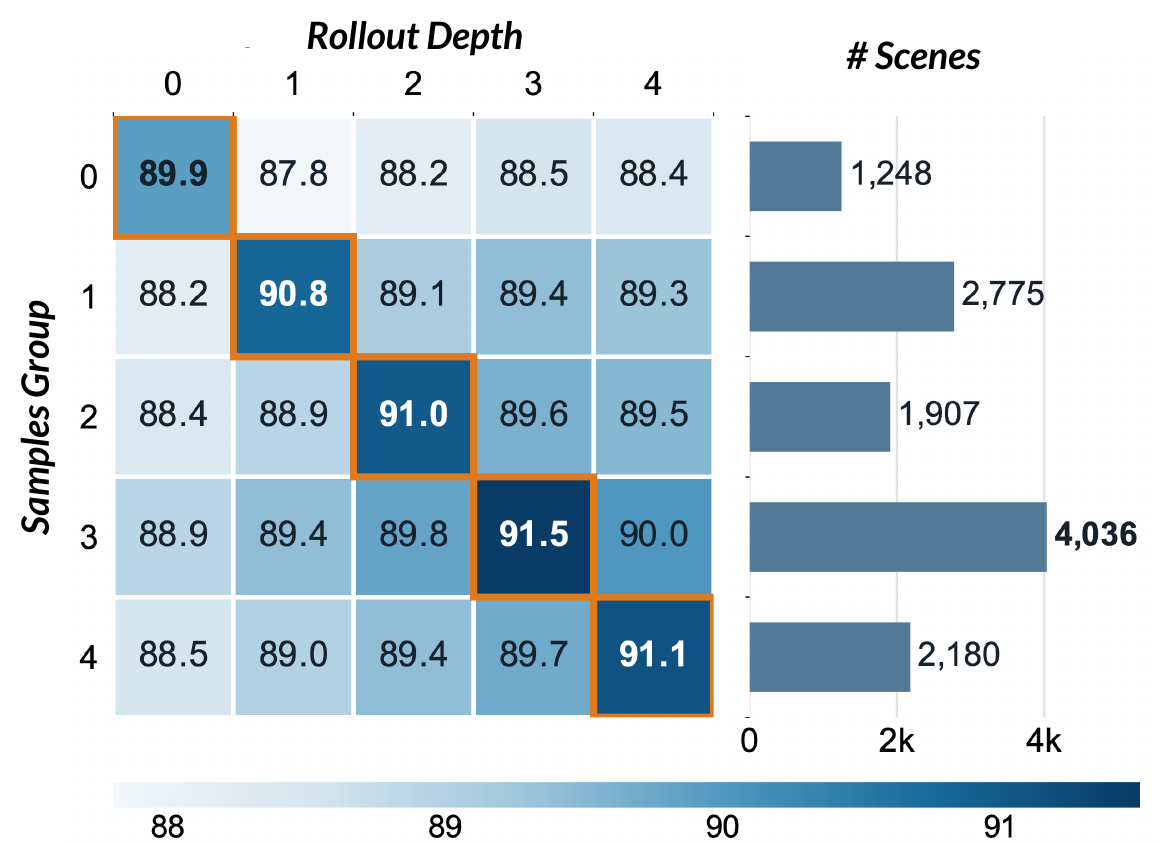}
    \caption{Performance under different rollout depths.}
    \label{fig:rollout}
\end{wrapfigure}

\begin{figure}[t]
    \centering
    \includegraphics[width=0.7\columnwidth]{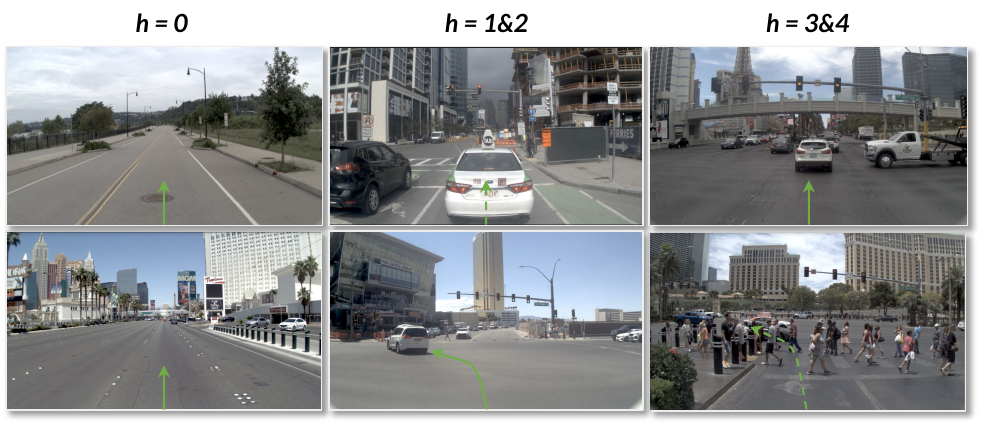}
    \caption{Representative scenes grouped by their preferred rollout depths. Scenes with denser traffic and more involved interactions generally benefit from additional rollout.} %
    \label{fig:scenes}
\end{figure}

To examine whether a single rollout depth can serve all scenes, we evaluate every sample at fixed depths $h\in\{0,1,2,3,4\}$ and group it by the depth $h^*$ yielding its highest planning score. Fig.~\ref{fig:rollout} shows substantial groups at every depth: 1,248 scenes perform best without rollout, whereas 4,036 and 2,180 scenes favor depths 3 and 4, respectively. For the $h^*=0$ group, increasing the depth from 0 to 4 reduces EPDMS from 89.9 to 88.4. Conversely, for the $h^*=4$ group, it improves from 88.5 to 91.1. Thus, deeper imagination benefits some scenes but is unnecessary or even detrimental to others. The examples in Fig.~\ref{fig:scenes} provide an intuitive view of this variation. The $h=0$ examples present open roads and limited immediate interaction, for which the observation already provides sufficient evidence for planning. The $h=1\&2$ examples contain traffic lights, vehicles, or road construction, introducing interactions that benefit from modest future context. The $h=3\&4$ cases involve dense pedestrian flow or crossing vehicles, where future evolution is ambiguous and additional prediction is valuable. These examples explain why the best-performing depth varies across scenes and why no fixed rollout budget is uniformly optimal.

\begin{figure*}[t]
    \centering
    \includegraphics[width=\linewidth]{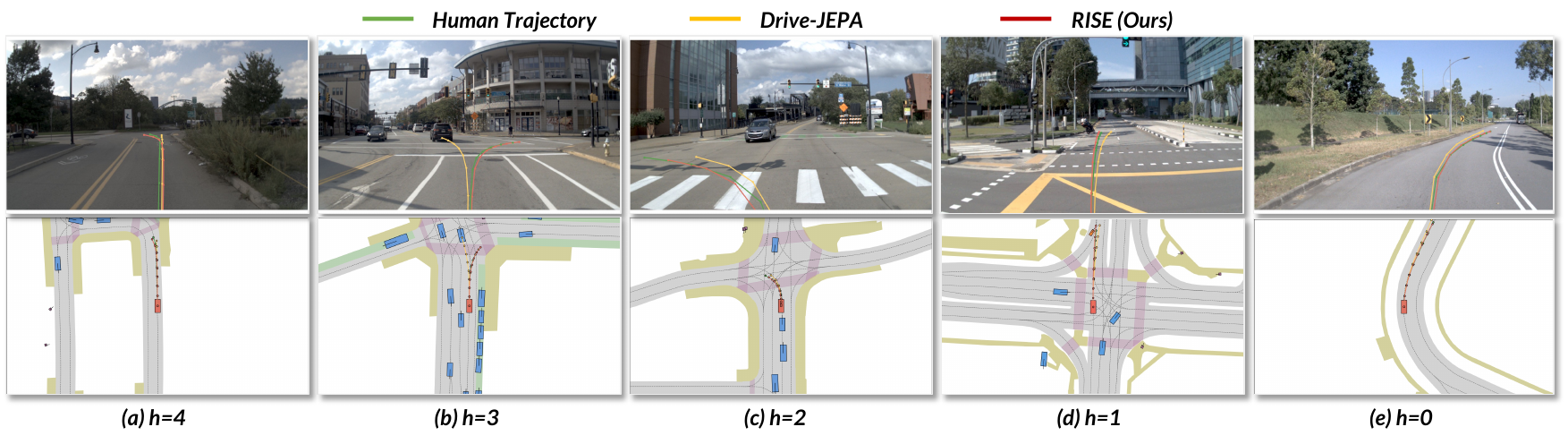}
    \caption{Qualitative trajectory comparison on NAVSIM at depths $h=4$ to $h=0$. The rows show front-camera views and corresponding bird's-eye-view scenes. Human, Drive-JEPA, and RISE trajectories are shown in green, yellow, and red, respectively. RISE remains aligned with human trajectories and feasible lane geometry across scene-dependent rollout depths.}
    \label{fig:vis}
\end{figure*}

\subsubsection{Does CounterDrive Improve Safety-Critical Evaluation?}
To assess whether CounterDrive improves safety-critical evaluation, we evaluate identical models with and without CounterDrive on the held-out counterfactual test set. As shown in Table~\ref{CounterDrive}, the model without CounterDrive remains close to random risk ranking, with AUC scores of 0.49--0.52 and an accident-recognition accuracy of 0.51. CounterDrive increases the four horizon-specific AUC scores to 0.93--0.96 and accuracy to 0.96, demonstrating substantially improved risk discrimination and accident awareness.

\begin{table}[h]
\centering
\small
\setlength{\tabcolsep}{8pt}
\begin{tabular}{c|cccc|c}
\toprule
\multirow{2}{*}{CounterDrive}& \multicolumn{4}{c|}{AUC$\uparrow$} & \multirow{2}{*}{Acc$\uparrow$}\\
\cline{2-5}
 & @1s & @2s & @3s & @4s &  \\
\hline
          & 0.49 & 0.52 & 0.51 & 0.50 & 0.51 \\
\checkmark & \textbf{0.96} & \textbf{0.95} & \textbf{0.96} & \textbf{0.93} & \textbf{0.96} \\
\bottomrule
\end{tabular}
\caption{Effect of CounterDrive on safety-critical evaluation. AUC@$h$ measures hazard--safe risk ranking at horizon $h$. Acc measures accident recognition.}
\label{CounterDrive}
\end{table}

\subsubsection{Can RISE Benefit Other Paradigm?}
To evaluate whether the proposed Scheduler generalizes beyond our base WAM, we reproduce DAWN, a representative World--Action Interactive Model, and integrate the Scheduler without modifying its Predictor or Planner. As shown in Table~\ref{DAWN}, the resulting model improves PDMS from 89.1 to 90.3. Consistent gains are also observed in NC, DAC, EP, and TTC, with particularly clear improvements of 2.7 points in EP and 2.3 points in TTC, while retaining a perfect collision score. These results suggest that the Scheduler can serve as a plug-in adaptive computation module for different world--action modeling paradigms, rather than being tied to a specific WAM.

\begin{table}[h]
\centering
\setlength{\tabcolsep}{4pt}
\begin{tabular}{l|cccccc}
\toprule
Method        & NC$\uparrow$  & DAC$\uparrow$ & EP$\uparrow$  & C$\uparrow$  & TTC$\uparrow$ & PDMS$\uparrow$ \\
\hline
DAWN   & 98.7 & 95.9 & 84.3 & 100 & 96.0 & 89.1  \\
+ Scheduler  & \textbf{99.9} & \textbf{96.8} &\textbf{ 87.0} & 100 & \textbf{98.3} & \textbf{90.3}  \\
\bottomrule
\end{tabular}
\caption{Transfer of the RISE Scheduler to DAWN. Adding the Scheduler consistently improves planning performance without modifying the underlying WAIM.}
\label{DAWN}
\end{table}

\subsubsection{Can RISE Generate Plausible and Safe Trajectories?}

Fig.~\ref{fig:vis} compares RISE with Drive-JEPA across representative scenes. In the intersection and turning cases of Fig.~\ref{fig:vis}(b)--(d), Drive-JEPA exhibits lateral or directional deviations from the local lane geometry, whereas RISE remains within feasible lane corridors and closely matches the human trajectory. In the curved-road cases of Fig.~\ref{fig:vis}(a) and (e), RISE likewise follows the reference closely, including at $h=0$, while Drive-JEPA shows greater lateral deviation. These examples show that RISE produces plausible, lane-consistent trajectories across varied road structures and selected rollout depths.

\section{Conclusion}
We presented RISE, a plug-in adaptive imagination framework that replaces fixed-depth WAM rollout with scene-dependent computation. Its Latent Evaluator estimates current risk and Future Planning Gain, allowing the Scheduler to make cost-aware sequential \textsc{Roll}/\textsc{Stop} decisions. We also introduced CounterDrive for counterfactual future learning and risk evaluation. Experiments on nuScenes and NAVSIM demonstrate competitive planning performance while balancing planning quality and inference cost.

\section{Limitations}
Although adaptive rollout is designed as a general principle for world action models, our experiments currently focus on autonomous driving, and its applicability to other domains remains to be explored. Furthermore, due to the cost of generating and filtering counterfactual samples, CounterDrive does not yet provide one-to-one coverage of the NAVSIM training set. Nevertheless, the primary goal of this work is to establish the necessity of scene-adaptive rollout, which is consistently supported by our experiments. Future work will extend the evaluation to broader domains and further scale the counterfactual dataset.

\section{Acknowledgments}

This work was supported in part by the Research and Application of Key Technologies for L4 End-to-End Autonomous Driving Based on Multi-modal Large Language Models under Grant 202423dl2050005, and in part by Research and Application of the Next-Generation General-Purpose Intelligent Robot Brain (Robo-GPT) under Grant 2024zd01.

\clearpage

\bibliographystyle{plainnat}
\bibliography{main}


\clearpage
\beginappendix
\section{Implementation Details}

\paragraph{Compute and data.} We implement RISE in PyTorch. The WAM and diffusion planners are trained with bfloat16 precision on eight NVIDIA A100 GPUs using a per-GPU batch size of 4, giving an effective batch size of 32. The lightweight Evaluator and Rollout Gate are trained offline in FP32 on one A100. We use four data-loading workers per process with pinned and persistent workers. Front-camera videos are sampled at 2\,Hz and resized to \(256\times512\). Each clip contains four observed frames. With a spatial patch size of 16 and a temporal tubelet size of 2, the four observations correspond to two observed latent steps. We set the maximum rollout horizons to \(H_{\mathrm{NAVSIM}}=4\) and \(H_{\mathrm{nuScenes}}=3\). Horizontal flipping, motion shifting, GridMask, AutoAugment, and random erasing are disabled.

\paragraph{World Action Model.} The visual encoder is a Drive-JEPA ViT-L/16 with causal attention. The action-conditioned Predictor contains 12 Transformer blocks with a hidden dimension of 384, 12 attention heads, rotary positional embeddings, and activation checkpointing. Each rollout step is conditioned on a three-dimensional ego-action vector. Ego-state conditioning, route-command conditioning, camera extrinsics, and parallel future prediction are disabled. A frozen Token-AE represents each latent frame using 128 tokens arranged on an \(8\times16\) latent grid. It uses a four-layer encoder, a two-layer decoder, and a two-layer causal temporal module with 16 attention heads and zero dropout. During Planner, Evaluator, and Scheduler training, the visual encoder and Predictor remain frozen.

\paragraph{Dynamic planners.} Both \(\Pi_0\) and \(\Pi_1\) use the same diffusion Transformer architecture with 12 layers, a hidden dimension of 384, 12 attention heads, an MLP ratio of 4, and zero dropout. The variance-preserving diffusion process uses \(\beta_{\min}=0.1\) and \(\beta_{\max}=20\). At inference, the Planner performs 20 denoising steps and generates \(K=6\) trajectory modes. Each pose is represented by \((x,y,\cos\theta,\sin\theta)\) with a temporal interval of \(0.5\,\mathrm{s}\). The classification, regression, velocity, and yaw losses are weighted by \(1.0\), \(1.0\), \(0.5\), and \(0.5\), respectively. The confidence temperature is 1.5, with classification and ignore thresholds of 2.0 and 0.2. The adaptive winner-take-all temperature starts at 8.0, decays with a factor of 0.984, and is lower-bounded by 0.1. During dynamic-prefix training, the full prefix \(h=H\) is sampled with probability 0.25, while the remaining probability is distributed uniformly over \(h\in\{0,\ldots,H-1\}\).

\paragraph{Planner optimization.} We optimize both planners using AdamW with \(\boldsymbol{\beta}=(0.9,0.999)\), \(\epsilon=10^{-8}\), and a weight decay of 0.04. For \(\Pi_0\), training lasts 20 epochs; the learning rate is linearly warmed up from \(2\times10^{-6}\) to \(2\times10^{-5}\) during the first two epochs and then cosine-decayed to zero. For \(\Pi_1\), training lasts 50 epochs; the learning rate is warmed up from \(10^{-5}\) to \(5\times10^{-5}\) during the first five epochs and then cosine-decayed to zero. We use deterministic execution and fixed diffusion noise for validation.

\paragraph{Evaluator and latent guidance.} The Evaluator first mean-pools the spatial tokens of each latent frame, applies LayerNorm, and encodes the prefix using a one-layer causal GRU with a hidden dimension of 512 and zero dropout. Two linear heads produce its task-evaluation profiles. The initial warm-up is performed for one epoch with 256 updates, a batch size of 4, and a learning rate of \(10^{-4}\). Real samples use a Huber loss with \(\delta=1\), while counterfactual hazard and trajectory-quality ranking losses use unit weights and a margin of 1.0. Local calibration uses one epoch with 256 updates, a batch size of 1, and a learning rate of \(5\times10^{-5}\). For each latent prefix, we sample four Gaussian perturbations with a scale of 0.05 and clip their per-token norm to 0.25; the local ordering loss has weight 1.0 and margin 0.1. The second profile head is subsequently calibrated for one epoch with a batch size of 1 and a learning rate of \(5\times10^{-5}\), while the shared prefix encoder is frozen. All Evaluator stages use AdamW with the same \(\boldsymbol{\beta}\), \(\epsilon\), and weight decay as the planners.

\paragraph{Rollout Gate.} The Rollout Gate applies LayerNorm followed by two 128-dimensional fully connected layers with GELU activations. It is distilled from complete horizon-wise utility curves for 50 epochs using AdamW, a batch size of 1, a constant learning rate of \(10^{-3}\), and a weight decay of 0.04. Its classification loss uses a temperature of 0.05, and the Smooth-L1 utility-regression term is weighted by 0.5. We define normalized rollout costs as \(c_h=h\) and train the conditional Gate with \(\lambda\in\{0,0.001,0.005,0.01,0.05\}\), using \(\lambda=0.005\) by default. Latent guidance performs two gradient steps on the latest imagined prefix with a step size of 0.05 and a maximum update norm of 0.25. The guided latent is detached before being passed to the Planner. Unless otherwise specified, all experiments use random seed 239.

\section{More Training Details of RISE}
\label{app:training_details}

\paragraph{Notation and conventions.} We use $H$ for the abstract maximum latent rollout horizon and $H_{\mathcal D}$ for its dataset-specific realization. In our experiments, $H_{\mathrm{NAVSIM}}=4$ and $H_{\mathrm{nuScenes}}=3$. We use $P$ for the number of future poses in a planned trajectory and $J$ for the number of Planner modes. The indices $h\in\{0,\ldots,H_{\mathcal D}\}$, $p\in\{1,\ldots,P\}$, and $s\in\{1,\ldots,J\}$ denote rollout depth, trajectory pose, and Planner mode, respectively. Index $i$ denotes a training sample or, when $i\in\mathcal I_{\mathrm{pair}}$, a verified factual--counterfactual source pair; $j$ denotes a valid continuation depth, and $m$ denotes a risk-refinement iteration. The real and counterfactual datasets are denoted by $\mathcal D_{\mathrm r}$ and $\mathcal D_{\mathrm c}$.

The Predictor, initial Planner, final Planner, Latent Evaluator, Rollout Gate, risk evaluator, and planning-score evaluator are denoted by $\mathcal P$, $\Pi_0$, $\Pi_1$, $\operatorname{Eval}_{\theta_E}$, $G_{\theta_G}$, $\mathcal E_{\mathrm{risk}}$, and $\mathcal E_{\mathrm{plan}}$, respectively. The relative-pose action mapping and token-pooling operator are denoted by $\mathcal A$ and $\rho$. We use $|\mathcal S|$ for the cardinality of a set $\mathcal S$, $\mathbb I[\cdot]$ for an indicator function, $[a]_+=\max(a,0)$, $\sigma(\cdot)$ for the logistic sigmoid, and $\operatorname{sg}(\cdot)$ for stop-gradient.

All scalar regression losses use the Huber function
\begin{equation}
\ell_\kappa(a,b)=
\begin{cases}
\frac{1}{2}(a-b)^2,
& |a-b|\leq\kappa,\\
\kappa\left(|a-b|-\frac{\kappa}{2}\right),
& \text{otherwise},
\end{cases}
\label{eq:app_huber}
\end{equation}
where $\kappa>0$ is the transition point. We use $\kappa=1$ in all experiments.

\subsection{Stage I: Predictor and Initial Planner}

\paragraph{Future-latent prediction.} For each sample, the frozen target Encoder produces the future latent target $z_{t+h}$. Starting from the observed latent, the Predictor recursively generates the maximum future prefix $\hat Z_{H_{\mathcal D}}$. The per-step latent loss is
\begin{equation}
\ell_z(\hat z,z)=
\frac{
 \left\|
 \operatorname{LN}(\hat z)
 -
 \operatorname{sg}(\operatorname{LN}(z))
 \right\|_1
}{
 N_{\mathrm{tok}}D_z
},
\label{eq:app_latent_loss}
\end{equation}
where $\operatorname{LN}$ denotes token-wise LayerNorm, $N_{\mathrm{tok}}$ is the number of tokens in one latent step, and $D_z$ is their embedding dimension. The Predictor objective is
\begin{equation}
\mathcal L_{\mathrm{Pre}}
=
\frac{1}{H_{\mathcal D}}
\sum_{h=1}^{H_{\mathcal D}}
\ell_z
\left(
\hat z_{t+h},
z_{t+h}
\right).
\label{eq:app_predictor_loss}
\end{equation}
This objective directly supervises the autoregressive rollout and is averaged over the minibatch. Real and accepted CounterDrive samples use the same loss. For CounterDrive, the generated future video is encoded into $z_{t+h}$, while relative pose changes between adjacent recovered ego poses provide the ego-motion conditioning. At inference, the same representation is computed only from adjacent poses in the observed ego-motion history; no future ground-truth pose is used. No route command is used.

\paragraph{Diffusion Planner objective.} At training time, we sample a diffusion time $u\sim\mathcal U(0,1)$ and Gaussian noise $\boldsymbol\epsilon\sim\mathcal N(0,\mathbf I)$. Under the variance-preserving schedule, the noisy trajectory representation is
\begin{equation}
\tau_i^{(u)}
=
\alpha_u\tau_i^*
+
\sigma_u\boldsymbol\epsilon,
\qquad
\alpha_u^2+\sigma_u^2=1.
\label{eq:app_vp_corruption}
\end{equation}
Conditioned on $\tau_i^{(u)}$, $u$, the observed latent, and the available future prefix, the Planner predicts clean trajectory candidates and their confidence logits. The losses below are evaluated on these denoised predictions, and the expectation over $u$ and $\boldsymbol\epsilon$ is approximated by minibatch sampling. The Planner additionally conditions on the observed ego-motion history and current ego kinematics; for compactness, these two inputs are suppressed in all subsequent $\Pi_0(\cdot)$ and $\Pi_1(\cdot)$ notation.

The Planner produces $J$ candidate trajectories. Let $\hat\tau_{i,s}$ denote candidate $s$ for sample $i$, and let $\tau_i^*$ denote the corresponding ground-truth trajectory. Each trajectory contains $P$ poses represented by $(x,y,\cos\theta,\sin\theta)$.

For a Planner minibatch of size $B_{\mathrm{pla}}$, the displacement error of mode $s$ is
\begin{equation}
d_{i,s}
=
\frac{1}{P}
\sum_{p=1}^{P}
\left\|
\hat{\mathbf p}_{i,s,p}
-
\mathbf p_{i,p}^*
\right\|_2,
\label{eq:app_mode_distance}
\end{equation}
where $\hat{\mathbf p}_{i,s,p}=(\hat x_{i,s,p},\hat y_{i,s,p})$ and $\mathbf p_{i,p}^*=(x_{i,p}^*,y_{i,p}^*)$.

The annealed winner-take-all weight is
\begin{equation}
\pi_{i,s}^{\mathrm{aw}}
=
\operatorname{sg}
\left(
\frac{
\exp(-d_{i,s}/T_{\mathrm{aw}})
}{
\sum_{s'=1}^{J}
\exp(-d_{i,s'}/T_{\mathrm{aw}})
}
\right),
\label{eq:app_awta}
\end{equation}
where $T_{\mathrm{aw}}>0$ is the annealed mode temperature. The position regression loss is
\begin{equation}
\begin{aligned}
\mathcal L_{\mathrm{xy}}
={}&
\frac{1}{B_{\mathrm{pla}}P}
\sum_{i=1}^{B_{\mathrm{pla}}}
\sum_{s=1}^{J}
\sum_{p=1}^{P}
\pi_{i,s}^{\mathrm{aw}}
\cdot
\left[
\ell_\kappa(\hat x_{i,s,p},x_{i,p}^*)
+
\ell_\kappa(\hat y_{i,s,p},y_{i,p}^*)
\right].
\end{aligned}
\label{eq:app_xy_loss}
\end{equation}

Let
\begin{equation}
\begin{aligned}
\hat{\mathbf y}_{i,s,p}
&=
(\widehat{\cos\theta}_{i,s,p},
 \widehat{\sin\theta}_{i,s,p}),\\
\mathbf y_{i,p}^*
&=
(\cos\theta_{i,p}^*,
 \sin\theta_{i,p}^*).
\end{aligned}
\end{equation}
The yaw consistency loss is
\begin{equation}
\begin{aligned}
\mathcal L_{\mathrm{yaw}}
={}&
\frac{1}{2B_{\mathrm{pla}}P}
\sum_{i=1}^{B_{\mathrm{pla}}}
\sum_{s=1}^{J}
\sum_{p=1}^{P}
\pi_{i,s}^{\mathrm{aw}}\cdot
\left(
1-
\frac{
\langle
\hat{\mathbf y}_{i,s,p},
\mathbf y_{i,p}^*
\rangle
}{
\|\hat{\mathbf y}_{i,s,p}\|_2
\|\mathbf y_{i,p}^*\|_2
}
\right).
\end{aligned}
\label{eq:app_yaw_loss}
\end{equation}

Let $g_{i,s}$ be the predicted confidence logit of mode $s$. Its soft target is
\begin{equation}
\pi_{i,s}^{\mathrm{mode}}
=
\operatorname{sg}
\left(
\frac{
\exp(-d_{i,s}/T_{\mathrm{conf}})
}{
\sum_{s'=1}^{J}
\exp(-d_{i,s'}/T_{\mathrm{conf}})
}
\right),
\label{eq:app_conf_target}
\end{equation}
where $T_{\mathrm{conf}}>0$ is the confidence temperature. Define
\begin{equation}
\begin{aligned}
d_i^\star &= \min_s d_{i,s},\\
s_i^\star &= \arg\min_s d_{i,s},\\
\mathcal I_{\mathrm{mode}}
&=
\{i:d_i^\star<\xi_{\mathrm{cls}}\},\\
\mathcal J_i
&=
\{s:d_{i,s}-d_i^\star>\xi_{\mathrm{ign}}\}
\cup\{s_i^\star\},
\end{aligned}
\end{equation}
where $\xi_{\mathrm{cls}}$ removes samples for which no mode is sufficiently close to the target, and $\xi_{\mathrm{ign}}$ removes ambiguous non-winning modes. The mode-confidence loss is
\begin{equation}
\begin{aligned}
\mathcal L_{\mathrm{mode}}
=
-\frac{1}{|\mathcal I_{\mathrm{mode}}|}
\sum_{i\in\mathcal I_{\mathrm{mode}}}
\sum_{s\in\mathcal J_i}
\pi_{i,s}^{\mathrm{mode}}
\log
\frac{\exp(g_{i,s})}
{\sum_{s'=1}^{J}\exp(g_{i,s'})}.
\end{aligned}
\label{eq:app_mode_loss}
\end{equation}
If $\mathcal I_{\mathrm{mode}}$ is empty, this term is set to zero. The complete trajectory objective is
\begin{equation}
\ell_\tau
=
\beta_{\mathrm{xy}}\mathcal L_{\mathrm{xy}}
+
\beta_{\mathrm{yaw}}\mathcal L_{\mathrm{yaw}}
+
\beta_{\mathrm{mode}}\mathcal L_{\mathrm{mode}},
\label{eq:app_trajectory_loss}
\end{equation}
where the expectation over sampled diffusion time and noise is implemented by the corruption process in Eq.~\eqref{eq:app_vp_corruption}. Since our trajectory representation contains no velocity channels, no velocity loss is used.

\paragraph{Initial variable-prefix Planner.} The Encoder and Predictor are frozen before Planner training. For each real sample, the Predictor produces all prefixes $\{\hat Z_h\}_{h=0}^{H_{\mathcal D}}$, where $\hat Z_h=\hat z_{t+1:t+h}$ and $\hat Z_0=\emptyset$. We sample $h$ uniformly and optimize
\begin{equation}
\mathcal L_{\mathrm{Pla}}^0
=
\ell_\tau
\left(
\Pi_0(z_t,\hat Z_h),
\tau^*
\right),
\qquad
h\sim\mathcal U\{0,\ldots,H_{\mathcal D}\}.
\label{eq:app_p0_loss}
\end{equation}
Only real samples provide Planner trajectory supervision.

\subsection{Stage II: Latent Evaluator and Guided Planner}

\paragraph{Geometry-based risk and planning scores.} For each Planner output, we select the candidate with the highest predicted confidence. On real data, the fixed risk evaluator returns
\begin{equation}
\begin{aligned}
r
&=
\mathcal E_{\mathrm{risk}}(\tau,\tau^*)\\
&=
\frac{
4R_{\mathrm{col}}
+
R_{\mathrm{near}}
+
R_{\mathrm{traj}}
+
0.2R_{\mathrm{comf}}
}{6.2},
\end{aligned}
\label{eq:app_risk_score}
\end{equation}
where $R_{\mathrm{col}}$, $R_{\mathrm{near}}$, $R_{\mathrm{traj}}$, and $R_{\mathrm{comf}}$ are the collision, near-miss, normalized trajectory-error, and comfort risks. Every component and the resulting risk lie in $[0,1]$, with larger values indicating less desirable trajectories. The corresponding higher-is-better planning score is
\begin{equation}
q
=
\mathcal E_{\mathrm{plan}}(\tau,\tau^*)
=
1-\mathcal E_{\mathrm{risk}}(\tau,\tau^*).
\label{eq:app_planning_score_definition}
\end{equation}
The trajectory-error normalization and geometric checks follow the native planning horizon and coordinate system of each dataset. Both evaluators are fixed, used only to construct targets, and never differentiated through.

\paragraph{Latent Evaluator.} Given the observed latent $z_t$ and prefix $\hat Z_h$, the Latent Evaluator predicts
\begin{equation}
\begin{aligned}
(\mathbf R_h,\mathbf B_h)
&=
\operatorname{Eval}_{\theta_E}(z_t,\hat Z_h),\\
\mathbf R_h
&=
[r_1,\ldots,r_h],\\
\mathbf B_h
&=
[b_{h\rightarrow h+1},\ldots,
b_{h\rightarrow H_{\mathcal D}}],
\end{aligned}
\label{eq:app_latent_evaluator}
\end{equation}
where $\mathbf R_h$ is the Risk Profile and $\mathbf B_h$ is the Future Planning Gain Profile. Each $r_k$ estimates the trajectory risk associated with the prefix ending at depth $k$, while $b_{h\rightarrow j}$ estimates the planning-score change from continuing to depth $j$ relative to stopping with the current prefix at depth $h$. For $h=0$, $\mathbf R_0$ is empty and $\mathbf B_0$ is predicted from the observed latent together with the learned empty-prefix embedding.

The Evaluator spatially averages the tokens of each latent step, applies LayerNorm, and processes the resulting sequence with a one-layer causal GRU. A risk projection produces the scalar estimate associated with each available prefix. A gain projection produces an $H_{\mathcal D}$-dimensional vector from the current recurrent state; entries for invalid continuation depths are masked. Consequently, all predictions depend only on the prefix available at the current rollout step.

\paragraph{Real Risk Profile regression.} For real sample $i$ and prefix ending at depth $k$, the target is
\begin{equation}
r_{i,k}^{*}
=
\mathcal E_{\mathrm{risk}}
\left(
\Pi_0(z_{i,t},\hat Z_{i,k}),
\tau_i^*
\right).
\label{eq:app_real_risk_target}
\end{equation}
For a real-data minibatch of size $B_{\mathrm r}$, we optimize
\begin{equation}
\mathcal L_{\mathrm{real}}
=
\frac{1}{B_{\mathrm r}H_{\mathcal D}}
\sum_{i=1}^{B_{\mathrm r}}
\sum_{k=1}^{H_{\mathcal D}}
\ell_\kappa
\left(
r_{i,k},
r_{i,k}^{*}
\right).
\label{eq:app_real_risk}
\end{equation}

\paragraph{Verified counterfactual risk ranking.} CounterDrive does not provide reliable future-agent geometry and therefore does not receive a scalar target from $\mathcal E_{\mathrm{risk}}$. Let $\mathcal I_{\mathrm{pair}}$ denote the verified factual--counterfactual source pairs for which the counterfactual future contains an annotated incident within the modeled horizon and is ranked as riskier than its factual source. Pairing is defined only within the selected source subset and does not imply one-to-one coverage of either original dataset.

For pair $i$, the annotated incident frame is temporally aligned with latent step $k_i^{\mathrm{inc}}$. We define
\begin{equation}
m_{i,k}
=
\mathbb I
\left[
k\geq k_i^{\mathrm{inc}}
\right].
\label{eq:app_incident_mask}
\end{equation}
Only risk estimates at and after the incident onset contribute to the ranking objective:
\begin{equation}
\mathcal L_{\mathrm{rank}}
=
\frac{1}{|\mathcal I_{\mathrm{pair}}|}
\sum_{i\in\mathcal I_{\mathrm{pair}}}
\frac{1}{\sum_{k=1}^{H_{\mathcal D}}m_{i,k}}
\sum_{k=1}^{H_{\mathcal D}}
m_{i,k}
\left[
\gamma
+
r_{i,k}^{\mathrm{fac}}
-
r_{i,k}^{\mathrm{cf}}
\right]_+ .
\label{eq:app_risk_rank}
\end{equation}
Thus, a hazardous counterfactual is assigned higher predicted risk than its factual source after the verified incident becomes observable. Real samples without a verified counterfactual partner are excluded only from this ranking term.

\paragraph{Temporally localized risk calibration.} The onset mask separates supervision before and after the incident becomes observable. While $\mathcal L_{\mathrm{rank}}$ enforces higher counterfactual risk at and after the onset, the local term aligns the paired risks beforehand:
\begin{equation}
\mathcal L_{\mathrm{loc}}
=
\frac{1}{|\mathcal I_{\mathrm{pair}}|}
\sum_{i\in\mathcal I_{\mathrm{pair}}}
\frac{
\sum_{k=1}^{H_{\mathcal D}}
(1-m_{i,k})
\ell_\kappa
\left(
r_{i,k}^{\mathrm{fac}},
r_{i,k}^{\mathrm{cf}}
\right)
}{
\max\left(
1,
\sum_{k=1}^{H_{\mathcal D}}(1-m_{i,k})
\right)
}.
\label{eq:app_local_risk}
\end{equation}
This prevents the counterfactual prefix from being labeled as riskier before the annotated incident is revealed and does not require geometry targets for generated CounterDrive clips. The complete Risk Profile objective is
\begin{equation}
\mathcal L_R
=
\mathcal L_{\mathrm{real}}
+
\beta_{\mathrm{cf}}
\mathcal L_{\mathrm{rank}}
+
\beta_{\mathrm{loc}}
\mathcal L_{\mathrm{loc}}.
\label{eq:app_complete_risk}
\end{equation}

\paragraph{Risk-guided latent refinement.} After Risk Profile training, the risk-prediction branch is frozen. For a non-empty prefix, let $\boldsymbol{\epsilon}_h^m$ be the refinement residual at iteration $m$, initialized by $\boldsymbol{\epsilon}_h^0=0$. Token-wise norm clipping is
\begin{equation}
\operatorname{clip}_{\varepsilon_{\mathrm r}}(\mathbf a)
=
\begin{cases}
\mathbf 0,
& \|\mathbf a\|_2=0,\\
\mathbf a
\min
\left(
1,
\frac{\varepsilon_{\mathrm r}}
{\|\mathbf a\|_2}
\right),
& \text{otherwise},
\end{cases}
\label{eq:app_risk_clip}
\end{equation}
where $\varepsilon_{\mathrm r}$ is the refinement trust radius. We reduce the accumulated prefix risk through
\begin{equation}
\begin{aligned}
\boldsymbol{\epsilon}_h^{m+1}
=
\operatorname{clip}_{\varepsilon_{\mathrm r}}
\Bigg(
\boldsymbol{\epsilon}_h^m
-
\eta
\nabla_{\boldsymbol{\epsilon}_h^m}
\sum_{k=1}^{h}
r_k
\left(
z_t,
\hat Z_h+\boldsymbol{\epsilon}_h^m
\right)
\Bigg).
\end{aligned}
\label{eq:app_risk_refinement}
\end{equation}
After $M_{\mathrm r}$ iterations, the risk-refined prefix is
\begin{equation}
\tilde Z_h
=
\hat Z_h+\boldsymbol{\epsilon}_h^{M_{\mathrm r}}.
\label{eq:app_refined_prefix}
\end{equation}
Refinement is skipped at $h=0$, for which $\tilde Z_0=\hat Z_0=\emptyset$.

\paragraph{Final Planner training.} We initialize $\Pi_1$ from $\Pi_0$ and train it on risk-refined real prefixes:
\begin{equation}
\mathcal L_{\mathrm{Pla}}^1
=
\ell_\tau
\left(
\Pi_1(z_t,\tilde Z_h),
\tau^*
\right),
\qquad
h\sim\mathcal U\{0,\ldots,H_{\mathcal D}\}.
\label{eq:app_p1_loss}
\end{equation}
The refined prefix is detached before entering $\Pi_1$. Therefore, $\mathcal L_{\mathrm{Pla}}^1$ updates only $\Pi_1$.

\paragraph{Future Planning Gain supervision.} For every real sample, we evaluate the final Planner at each valid rollout depth:
\begin{equation}
q_{i,h}
=
\mathcal E_{\mathrm{plan}}
\left(
\Pi_1(z_{i,t},\tilde Z_{i,h}),
\tau_i^*
\right).
\label{eq:app_planning_score}
\end{equation}
The realized Future Planning Gain Profile at depth $h$ is
\begin{equation}
\mathbf B_{i,h}^{*}
=
\left[
q_{i,h+1}-q_{i,h},
\ldots,
q_{i,H_{\mathcal D}}-q_{i,h}
\right].
\label{eq:app_gain_target}
\end{equation}
The gain branch is trained only on valid continuation entries:
\begin{equation}
\mathcal L_B
=
\frac{1}{B_{\mathrm r}}
\sum_{i=1}^{B_{\mathrm r}}
\sum_{h=0}^{H_{\mathcal D}-1}
\frac{1}{H_{\mathcal D}-h}
\sum_{j=h+1}^{H_{\mathcal D}}
\ell_\kappa
\left(
b_{i,h\rightarrow j},
q_{i,j}-q_{i,h}
\right).
\label{eq:app_gain_loss}
\end{equation}
These targets compare every valid continuation with the planning score already available from the current prefix; they do not supervise a final rollout depth directly.

\subsection{Stage III: Rollout Gate}

\paragraph{Cost-adjusted continuation target.} Gate supervision uses real samples and their all-horizon planning scores. Given cumulative rollout cost $c_h$ and computation preference $\lambda\geq0$, the best remaining cost-adjusted gain from depth $h$ is
\begin{equation}
g_{i,h}^{*}
=
\max_{j\in\{h+1,\ldots,H_{\mathcal D}\}}
\left[
(q_{i,j}-q_{i,h})
-
\lambda(c_j-c_h)
\right].
\label{eq:app_gate_gain}
\end{equation}
The binary continuation target is
\begin{equation}
y_{i,h}^{*}
=
\mathbb I
\left[
g_{i,h}^{*}>0
\right],
\qquad
0\leq h<H_{\mathcal D}.
\label{eq:app_gate_label}
\end{equation}
A tie is assigned to \textsc{Stop}, and no Gate target is constructed at $h=H_{\mathcal D}$ because rollout terminates by construction.

\paragraph{Online Gate input.} For one sample and $0\leq h<H_{\mathcal D}$, the pooled latent feature and rollout-state feature are
\begin{equation}
\begin{aligned}
e_h
&=
[\rho(z_t),\rho(\hat Z_h)],\\
\xi_h
&=
[
\overline{\mathbf R}_h,
\overline{\mathbf B}_h,
h/H_{\mathcal D},
c_h,
\Delta c_{h+1}
],
\end{aligned}
\label{eq:app_gate_input}
\end{equation}
where $\rho$ denotes token pooling, $\rho(\emptyset)$ is a learned empty-prefix embedding, $\overline{\mathbf R}_h$ and $\overline{\mathbf B}_h$ are zero-padded to $H_{\mathcal D}$ entries, and $\Delta c_{h+1}=c_{h+1}-c_h$. Given $\lambda$, the Gate predicts
\begin{equation}
x_h
=
G_{\theta_G}(e_h,\xi_h,\lambda).
\label{eq:app_gate_output}
\end{equation}
The sign of $x_h$ indicates whether the predicted planning gain available from continued rollout justifies the additional cost.

\paragraph{Gate objective.} Let $\Lambda$ be the set of computation preferences used for Gate training and define
\begin{equation}
\Omega_G
=
\left\{
(i,h,\lambda):
1\leq i\leq B_G,\;
0\leq h<H_{\mathcal D},\;
\lambda\in\Lambda
\right\}.
\end{equation}
With $N_G=|\Omega_G|$, the Gate objective is
\begin{equation}
\mathcal L_G
=
\frac{1}{N_G}
\sum_{(i,h,\lambda)\in\Omega_G}
\ell_{\mathrm{BCE}}
\left(
\sigma(x_{i,h}),
y_{i,h}^{*}
\right).
\label{eq:app_gate_loss}
\end{equation}
Full-horizon enumeration is used only to construct training targets. At inference, the Gate observes only the currently available prefix, appends exactly one latent after each \textsc{Roll} decision, and reevaluates the extended prefix.

\paragraph{Hyperparameters and optimization.} We use $P=8$ trajectory poses, $J=6$ Planner modes, and $M_{\mathrm r}=2$ risk-refinement iterations. The Planner weights are $\beta_{\mathrm{xy}}=1$, $\beta_{\mathrm{yaw}}=0.5$, and $\beta_{\mathrm{mode}}=1$. We use $T_{\mathrm{conf}}=1.5$, $\xi_{\mathrm{cls}}=2.0$, and $\xi_{\mathrm{ign}}=0.2$. The aWTA temperature starts from $8.0$, is multiplied by $0.984$ after each epoch, and is lower-bounded by $0.1$.

For Risk Profile training, we use $\gamma=1$, $\beta_{\mathrm{cf}}=1$, and $\beta_{\mathrm{loc}}=1$. Risk refinement uses $\eta=0.05$ and $\varepsilon_{\mathrm r}=0.25$. For Gate training, $c_h=h$ and
\begin{equation}
\Lambda=\{0,0.001,0.005,0.01,0.05\}.
\end{equation}
The default inference preference is $\lambda=0.005$.


All experiments use a per-GPU batch size of 4. The Predictor is trained for 80 epochs with learning rate $2\times10^{-4}$. The initial Planner $\Pi_0$ is trained for 20 epochs with learning rate $2\times10^{-5}$, and $\Pi_1$ is trained for 50 epochs with learning rate $5\times10^{-5}$. The Latent Evaluator and Rollout Gate are trained for 50 epochs with learning rate $10^{-3}$. All stages use AdamW, weight decay $0.04$, and bfloat16 precision.

CounterDrive participates in Predictor training and, only for verified factual--counterfactual source pairs, in $\mathcal L_{\mathrm{rank}}$ and $\mathcal L_{\mathrm{loc}}$. Unpaired factual samples remain available for the standard real-data objectives. Final Planner, Future Planning Gain, and Rollout Gate supervision use real planning outcomes.

\clearpage



\section{More Experimental Results}
\label{app:more_ablations}

\subsection{How Does Rollout Depth Affect Different Scenes?}
\begin{figure}[h]
    \centering
    \includegraphics[width=0.85\columnwidth]{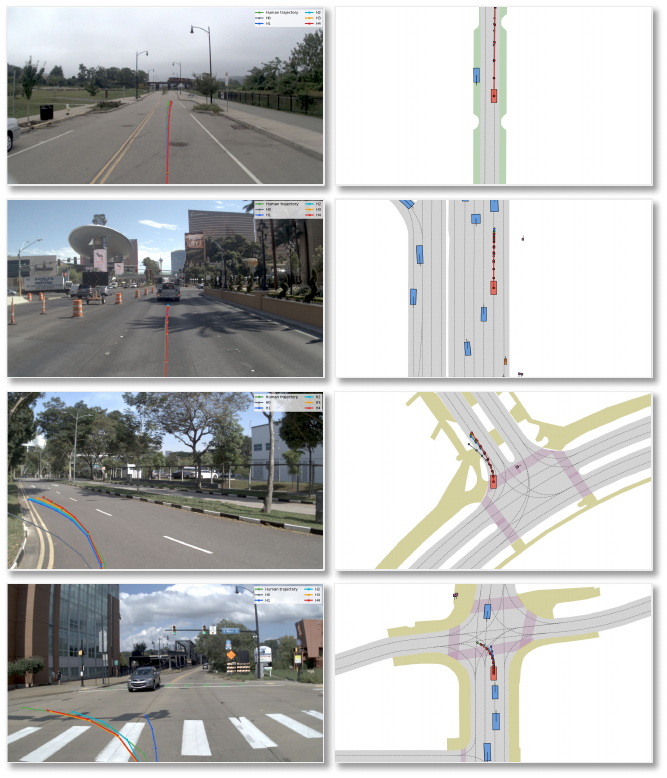}
    \caption{Trajectory predictions at different fixed rollout depths. Predictions remain nearly identical in simple scenes but vary noticeably in turning and intersection cases.}
    \label{fig:app_vis}
\end{figure}

Figure~\ref{fig:app_vis} compares trajectories generated at different fixed rollout depths. In the first two simple lane-following scenes, predictions at all depths nearly overlap, indicating that additional rollout has little influence when the current observation already supports a stable plan. In contrast, the turning and intersection cases exhibit clear differences in trajectory curvature and direction across rollout depths. This comparison shows that sensitivity to rollout depth is scene-dependent: simple scenes remain stable across depths, whereas complex scenes require an appropriate rollout budget.

\subsection{Latent Evaluator and Rollout Gate}

\begin{table}[h]
\centering
\caption{Ablation of the Latent Evaluator and Rollout Gate on NAVSIM with all other components fixed.}
\label{tab:app_evaluator_gate}
\setlength{\tabcolsep}{5pt}
\begin{tabular}{cc|cc}
\toprule
\textbf{Latent Evaluator} & \textbf{Gate} & \textbf{EPDMS$\uparrow$} & \textbf{PDMS$\uparrow$} \\
\midrule
 &  & 89.8 & 90.5 \\
 & \checkmark & 90.6 & 91.2 \\
\checkmark &  & 90.3 & 90.9 \\
\checkmark & \checkmark & \textbf{90.8} & \textbf{91.5} \\
\bottomrule
\end{tabular}
\end{table}

Table~\ref{tab:app_evaluator_gate} separates the contributions of the two Scheduler components. Adding the Latent Evaluator alone improves EPDMS/PDMS from 89.8/90.5 to 90.3/90.9, while the Gate alone raises them to 90.6/91.2. Combining both components gives the best result of 90.8/91.5, showing that evaluating predicted futures and adaptively allocating rollout provide complementary improvements.

\subsection{Variable-Prefix Planner Training Distribution}

\begin{table}[h]
\centering
\caption{Ablation of prefix-depth sampling for variable-prefix Planner training on NAVSIM. The distributions correspond to $h=0,\ldots,4$.}
\label{tab:app_prefix_distribution}
\setlength{\tabcolsep}{10pt}
\begin{tabular}{lcc}
\toprule
\textbf{Strategy} & \textbf{Training Distribution} & \textbf{EPDMS$\uparrow$} \\
\midrule
Uniform & $[0.20,0.20,0.20,0.20,0.20]$ & \textbf{89.8} \\
Extremes & $[0.50,0,0,0,0.50]$ & 87.4 \\
Short-heavy & $[0.225,0.225,0.225,0.225,0.10]$ & 89.6 \\
No-full & $[0.25,0.25,0.25,0.25,0]$ & 89.2 \\
\bottomrule
\end{tabular}
\end{table}

Table~\ref{tab:app_prefix_distribution} examines how the sampled prefix distribution affects variable-prefix Planner training. Uniform sampling performs best, while training only on the two extreme depths causes a substantial drop to 87.4 EPDMS. Short-heavy sampling remains competitive, but excluding the full prefix also degrades performance, indicating that balanced coverage of all valid depths is important.

\subsection{Number of Risk-Refinement Steps}

\begin{table}[h]
\centering
\caption{Effect of the number of risk-refinement steps on NAVSIM.}
\label{tab:app_refinement_steps}
\setlength{\tabcolsep}{10pt}
\begin{tabular}{ccc}
\toprule
\textbf{Steps $m$} & \textbf{PDMS$\uparrow$} & \textbf{EPDMS$\uparrow$} \\
\midrule
0 & 90.5 & 89.8 \\
1 & 90.7 & 90.1 \\
2 & \textbf{90.9} & \textbf{90.2} \\
4 & \textbf{90.9} & \textbf{90.2} \\
8 & \textbf{90.9} & \textbf{90.2} \\
\bottomrule
\end{tabular}
\end{table}

As shown in Table~\ref{tab:app_refinement_steps}, one refinement step already improves both metrics, and two steps increase PDMS/EPDMS from 90.5/89.8 to 90.9/90.2. Increasing $m$ to 4 or 8 provides no further gain, so we use two steps to avoid redundant optimization.

\clearpage

\subsection{Effect of the Computation-Cost Weight}

\begin{table}[h]
\centering
\caption{Effect of the computation-cost weight $\lambda$ on NAVSIM planning performance.}
\label{tab:app_lambda}
\setlength{\tabcolsep}{12pt}
\begin{tabular}{cc}
\toprule
\textbf{$\lambda$} & \textbf{EPDMS$\uparrow$} \\
\midrule
0 & 90.0 \\
0.001 & 90.7 \\
0.005 & \textbf{90.8} \\
0.01 & 88.6 \\
0.05 & 88.4 \\
\bottomrule
\end{tabular}
\end{table}

Table~\ref{tab:app_lambda} shows that a modest computation penalty improves EPDMS from 90.0 at $\lambda=0$ to 90.8 at $\lambda=0.005$. Larger values substantially reduce planning performance, consistent with over-penalizing continued rollout. We therefore use $\lambda=0.005$ as the default setting.

\clearpage

\section{Samples of CounterDrive}


\begin{figure}[h]
    \centering
    \includegraphics[width=1\columnwidth]{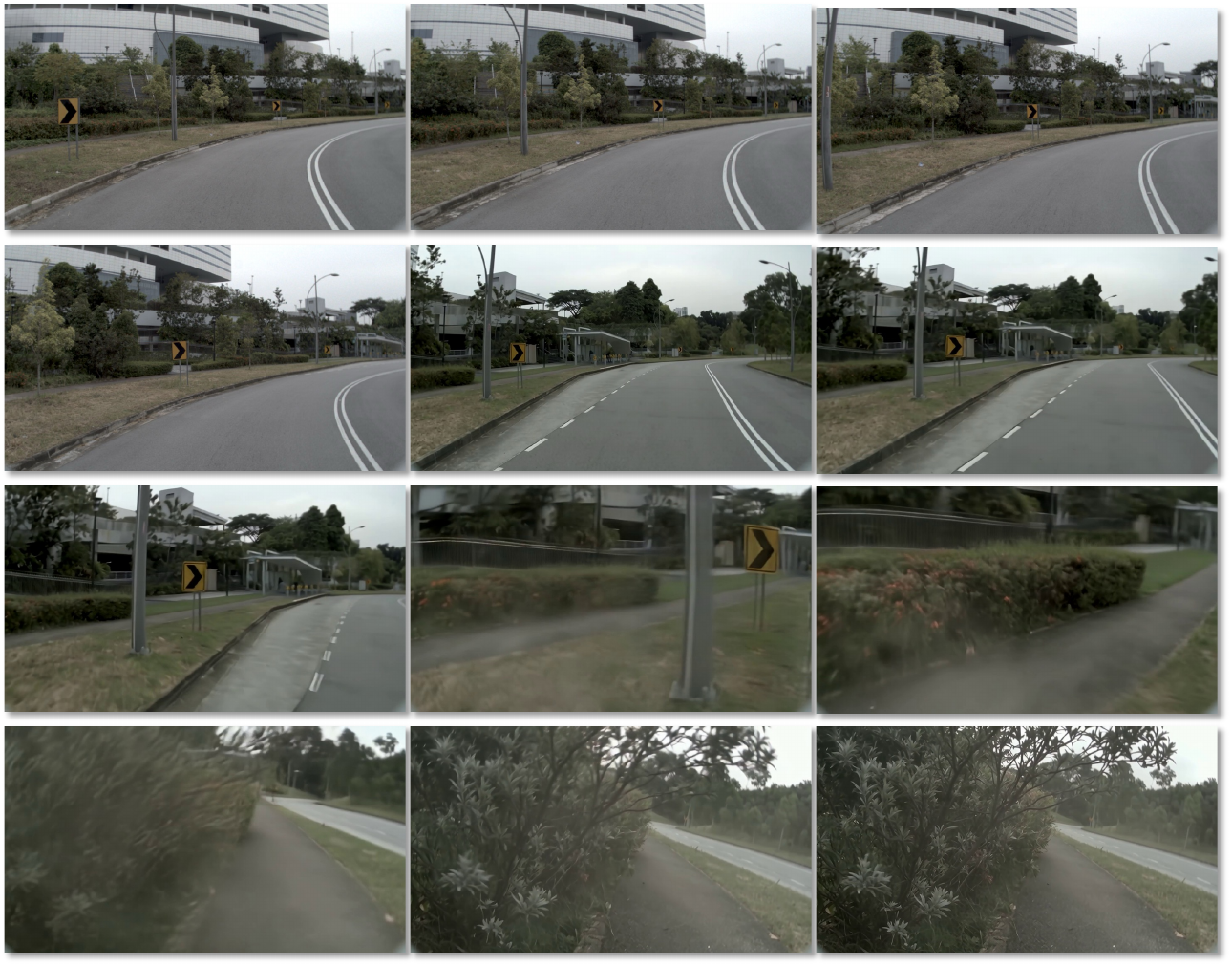}
    \caption{Sample 1 of CounterDrive.}
    \label{fig:cf1}
\end{figure}

\clearpage

\begin{figure}[h]
    \centering
    \includegraphics[width=1\columnwidth]{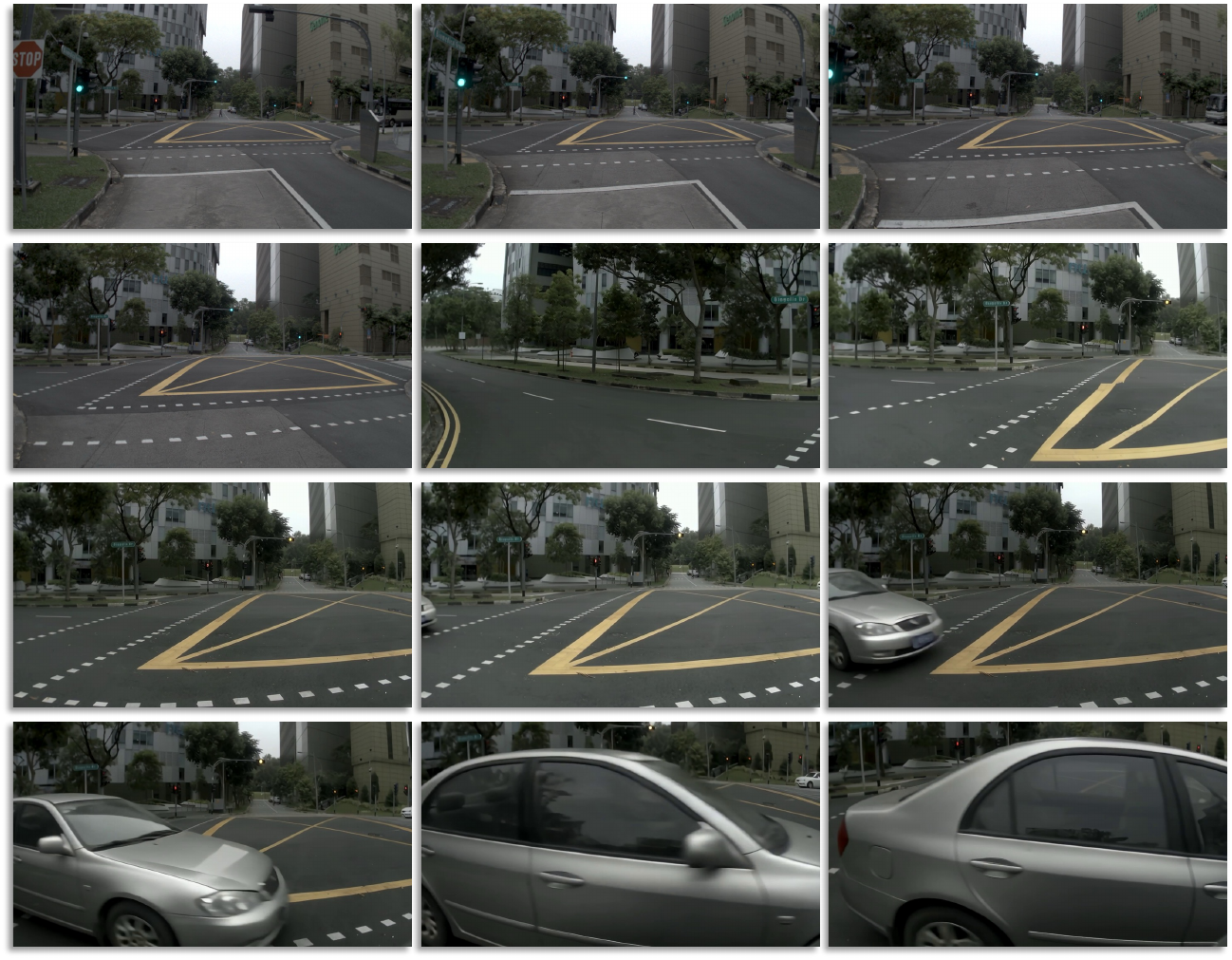}
    \caption{Sample 2 of CounterDrive.}
    \label{fig:cf2}
\end{figure}

\clearpage

\section{Prompts for CounterDrive Generation}
\label{app:prompt_construction}

CounterDrive constructs each video-generation prompt in two
steps. First, a vision-language model (VLM) describes the
near-field content of the source key frame and generates a
one-sentence accident description. Second, these two variable
components are concatenated with a manually written camera
constraint prompt and supplied to Wan together with the source
key frame. The operational prompts are written in Chinese; we
provide faithful English translations below.

\subsection{VLM-Based Prompt Construction}

Table~\ref{tab:vlm_prompt} presents the instructions used to
generate the two variable prompt components. The key-frame
description is restricted to nearby visible objects so that the
subsequent accident description remains grounded in the source
scene. The accident description explicitly specifies both the
relative accident location and the involved object.

\begin{table*}[h]
\centering
\caption{VLM instructions for constructing the variable components
of the Wan prompt.}
\label{tab:vlm_prompt}
\setlength{\tabcolsep}{7pt}
\begin{tabularx}{\textwidth}{
>{\raggedright\arraybackslash}p{0.19\textwidth} Y}
\toprule
\rowcolor{PromptHeader}
\textbf{Component} & \textbf{Prompt content and constraints} \\
\midrule

\rowcolor{PromptSection}
\multicolumn{2}{l}{
\textbf{A. Near-field key-frame description}
} \\

\textbf{VLM input}
&
One front-facing dashcam key frame. \\

\rowcolor{PromptLight}
\textbf{Role}
&
You are a near-field object-description assistant for dashcam
images. Given a front-facing dashcam image, describe only objects
immediately surrounding the camera in exactly one sentence. Do
not analyze accidents and do not describe the ego vehicle. \\

\textbf{Spatial range}
&
Only consider objects within approximately 30\,m in front of the
camera and 20\,m on either side of the road. \\

\rowcolor{PromptLight}
\textbf{Include}
&
Include vehicles immediately ahead using their vehicle type only,
such as ``double-decker bus'' or ``SUV.'' Also include nearby
roadside objects, such as trees, guardrails, building facades,
streetlights, or traffic signs. Mention the nearby road surface
only when necessary. \\

\textbf{Exclude}
&
Do not mention city names, street names, distant landmarks,
distant buildings, weather, illumination, time of day, detailed
traffic conditions, or inferred geographical locations. \\

\rowcolor{PromptLight}
\textbf{Output format}
&
Return exactly one plain sentence. Do not output JSON, Markdown,
numbering, explanations, or additional analysis. \\

\textbf{Example output}
&
``An SUV is immediately ahead, and a sedan is traveling in the
adjacent left lane.'' \\

\midrule
\rowcolor{PromptSection}
\multicolumn{2}{l}{
\textbf{B. Accident description}
} \\

\textbf{VLM input}
&
The source key frame and its near-field description generated in
Part A. \\

\rowcolor{PromptLight}
\textbf{Role}
&
You are an accident-event description assistant for dashcam
scenes. Based on the visible nearby objects, describe one
physically plausible accident that occurs immediately after the
source frame. \\

\textbf{Grounding}
&
Select an involved object that is visible in the key frame and
present in the near-field description. Do not introduce a new
vehicle, pedestrian, obstacle, or road structure. \\

\rowcolor{PromptLight}
\textbf{Required content}
&
Explicitly state the accident location relative to the ego
vehicle, such as front, front-left, or front-right, and identify
the involved object. Describe the event directly rather than
analyzing its cause. \\

\textbf{Temporal constraint}
&
The accident must be a continuous and plausible evolution from
the source key frame. Do not introduce camera cuts, viewpoint
changes, or discontinuous object motion. \\

\rowcolor{PromptLight}
\textbf{Output format}
&
Return exactly one plain sentence. Do not provide reasoning,
warnings, explanations, JSON, Markdown, or multiple candidate
events. \\

\textbf{Example output}
&
``The ego vehicle directly rear-ends the SUV ahead.'' \\

\bottomrule
\end{tabularx}
\end{table*}

\clearpage

\subsection{Wan Video-Generation Prompt}

The final Wan prompt contains three ordered components:
a fixed camera prompt, the VLM-generated key-frame description,
and the VLM-generated accident description. No additional
language-model rewriting is applied after concatenation.
Table~\ref{tab:wan_prompt} shows the template and an example.

\begin{table*}[h]
\centering
\caption{Construction of the prompt supplied to Wan for
counterfactual video generation.}
\label{tab:wan_prompt}
\setlength{\tabcolsep}{7pt}
\begin{tabularx}{\textwidth}{
>{\raggedright\arraybackslash}p{0.19\textwidth} Y}
\toprule
\rowcolor{PromptHeader}
\textbf{Component} & \textbf{Content} \\
\midrule

\textbf{Wan input}
&
The source key frame and the concatenated text prompt described
below. \\

\rowcolor{PromptLight}
\textbf{Fixed system prompt}
&
``Dashcam perspective with fixed camera height and pitch. The
camera height must remain unchanged, and no abrupt camera cuts
may occur. The video must remain in a first-person driving view,
and the ego vehicle must never appear in the image.'' \\

\textbf{Key-frame description}
&
A one-sentence near-field scene description generated by the VLM
using Part A of Table~\ref{tab:vlm_prompt}. \\

\rowcolor{PromptLight}
\textbf{Accident description}
&
A one-sentence accident description generated by the VLM using
Part B of Table~\ref{tab:vlm_prompt}. The sentence specifies the
accident location and involved object. \\

\textbf{Concatenation rule}
&
\texttt{<FIXED CAMERA PROMPT> + <KEY-FRAME DESCRIPTION> +}
\newline
\texttt{<ACCIDENT DESCRIPTION>} \\

\rowcolor{PromptSection}
\textbf{Complete example}
&
``Dashcam perspective with fixed camera height and pitch. The
camera height must remain unchanged, and no abrupt camera cuts
may occur. The video must remain in a first-person driving view,
and the ego vehicle must never appear in the image. An SUV is
immediately ahead, and a sedan is traveling in the adjacent left
lane. The ego vehicle directly rear-ends the SUV ahead.'' \\

\textbf{Wan output}
&
A temporally continuous counterfactual driving video that
preserves the source viewpoint and scene layout while realizing
the specified accident. \\

\bottomrule
\end{tabularx}
\end{table*}

\clearpage

\section{Algorithms of RISE}
\label{app:algorithms}

\subsection{Inference}

Here, $\operatorname{RiskRefine}$ denotes the iterative update in Eq.~\eqref{eq:app_risk_refinement} using the frozen Risk Profile branch of the Latent Evaluator; it is not a separate module.

\begin{algorithm}[h]
\caption{Adaptive inference with RISE}
\label{alg:rise_inference}
\begin{algorithmic}[1]
\Require Observation $O_t$, observed ego-pose history $\tau^{\mathrm{obs}}_{t-L:t}$, current ego kinematics $\mathbf s_t^{\mathrm{ego}}$, maximum horizon $H_{\mathcal D}$, computation preference $\lambda$
\Ensure Planned trajectory $\hat\tau$
\State $z_t\gets\operatorname{Encoder}(O_t)$
\State $a^{\mathrm{obs}}\gets \{\mathcal A(\tau^{\mathrm{obs}}_u, \tau^{\mathrm{obs}}_{u+1})\}_{u=t-L}^{t-1}$
\State $\hat Z_0\gets\emptyset$, $h\gets0$
\While{$h<H_{\mathcal D}$}
    \State $(\mathbf R_h,\mathbf B_h) \gets\operatorname{Eval}_{\theta_E}(z_t,\hat Z_h)$
    \State $e_h\gets[\rho(z_t),\rho(\hat Z_h)]$
    \State $\xi_h\gets [\overline{\mathbf R}_h,\overline{\mathbf B}_h, h/H_{\mathcal D},c_h,\Delta c_{h+1}]$
    \State $x_h\gets G_{\theta_G}(e_h,\xi_h,\lambda)$
    \If{$x_h\leq0$}
        \State \textbf{break}
    \EndIf
    \State $\hat z_{t+h+1} \gets\mathcal P(z_t,\hat Z_h,a^{\mathrm{obs}})$
    \State $\hat Z_{h+1} \gets\hat Z_h\cup\{\hat z_{t+h+1}\}$
    \State $h\gets h+1$
\EndWhile
\If{$h>0$}
    \State $\widetilde Z_h\gets \operatorname{RiskRefine} (z_t,\hat Z_h;\operatorname{Eval}_{\theta_E})$
\Else
    \State $\widetilde Z_h\gets\emptyset$
\EndIf
\State $(\{\hat\tau_s\},\{g_s\}) \gets\Pi_1(z_t,\widetilde Z_h,a^{\mathrm{obs}}, \mathbf s_t^{\mathrm{ego}})$
\State $\hat\tau\gets\hat\tau_{\arg\max_s g_s}$
\State \Return $\hat\tau$
\end{algorithmic}
\end{algorithm}

\subsection{Training}

\begin{algorithm*}[h]
\caption{Three-stage training procedure of RISE}
\label{alg:rise_training}
\begin{algorithmic}[1]
\Require Real data $\mathcal D_{\mathrm r}$, accepted CounterDrive data $\mathcal D_{\mathrm c}$, verified pair set $\mathcal I_{\mathrm{pair}}$, maximum horizon $H_{\mathcal D}$, and computation preferences $\Lambda$
\Ensure Predictor $\mathcal P$, Planners $\Pi_0,\Pi_1$, Latent Evaluator $\operatorname{Eval}_{\theta_E}$, and Rollout Gate $G_{\theta_G}$

\Statex \textbf{Stage I: Predictor and initial Planner}
\State Train $\mathcal P$ on $\mathcal D_{\mathrm r}\cup\mathcal D_{\mathrm c}$ using $\mathcal L_{\mathrm{Pre}}$
\State Freeze the Encoder and $\mathcal P$
\For{each sample in $\mathcal D_{\mathrm r}$}
    \State Generate $\{\hat Z_h\}_{h=0}^{H_{\mathcal D}}$ and sample $h\sim\mathcal U\{0,\ldots,H_{\mathcal D}\}$
    \State Update $\Pi_0(z_t,\hat Z_h)$ using $\mathcal L_{\mathrm{Pla}}^{0}$
\EndFor
\State Freeze $\Pi_0$

\Statex \textbf{Stage II: Latent Evaluator and final Planner}
\For{each real-data batch}
    \State Compute $r_{i,k}^{*} =\mathcal E_{\mathrm{risk}} (\Pi_0(z_{i,t},\hat Z_{i,k}),\tau_i^*)$
    \State Accumulate $\mathcal L_{\mathrm{real}}$
\EndFor
\For{each verified pair in $\mathcal I_{\mathrm{pair}}$}
    \State Map the incident onset to $k_i^{\mathrm{inc}}$ and construct $m_{i,k}$
    \State Accumulate $\mathcal L_{\mathrm{rank}}$ and $\mathcal L_{\mathrm{loc}}$
\EndFor
\State Update the Risk Profile branch using $\mathcal L_R$ and then freeze it
\State Initialize $\Pi_1\gets\Pi_0$
\For{each real sample}
    \State Sample $h$ and compute the risk-refined prefix $\widetilde Z_h$
    \State Update $\Pi_1(z_t,\widetilde Z_h)$ using $\mathcal L_{\mathrm{Pla}}^{1}$
\EndFor
\State Freeze $\Pi_1$
\For{each real sample}
    \State Enumerate $h=0,\ldots,H_{\mathcal D}$ and compute $q_h=\mathcal E_{\mathrm{plan}} (\Pi_1(z_t,\widetilde Z_h),\tau^*)$
    \For{$h=0,\ldots,H_{\mathcal D}-1$}
        \State Construct $\mathbf B_h^* =[q_{h+1}-q_h,\ldots,q_{H_{\mathcal D}}-q_h]$
    \EndFor
    \State Update the Future Planning Gain branch using $\mathcal L_B$
\EndFor

\Statex \textbf{Stage III: Rollout Gate}
\State Freeze $\mathcal P$, $\Pi_1$, and $\operatorname{Eval}_{\theta_E}$
\For{each real sample}
    \State Reuse the all-horizon planning scores $\{q_h\}$
    \ForAll{$\lambda\in\Lambda$ and $h=0,\ldots,H_{\mathcal D}-1$}
        \State $g_h^*\gets \max_{j>h}[(q_j-q_h)-\lambda(c_j-c_h)]$
        \State $y_h^*\gets\mathbb I[g_h^*>0]$
        \State Construct $(e_h,\xi_h)$ from the current prefix and predict $x_h=G_{\theta_G}(e_h,\xi_h,\lambda)$
        \State Update $G_{\theta_G}$ using $\ell_{\mathrm{BCE}}(\sigma(x_h),y_h^*)$
    \EndFor
\EndFor
\State \Return $\mathcal P,\Pi_0,\Pi_1, \operatorname{Eval}_{\theta_E},G_{\theta_G}$
\end{algorithmic}
\end{algorithm*}




\end{document}